\documentclass[preprint]{elsarticle}

\usepackage{geometry}
\usepackage{amsmath}
\usepackage{amsfonts}
\usepackage{booktabs}
\usepackage{pdfpages}
\usepackage{algorithm}
\usepackage{algpseudocode}
\usepackage{graphicx}
\usepackage{hyperref}
\usepackage{bm}

\usepackage{caption}
\usepackage{subcaption}
\usepackage{booktabs}
\usepackage{multirow}
\usepackage{adjustbox}  

\usepackage{booktabs}
\usepackage{pifont}
\newcommand{\cmark}{\ding{51}}
\newcommand{\xmark}{\ding{55}}

\journal{Journal of Intelligent Manufacturing}

\begin{document}

\begin{frontmatter}

\title{Learning to Hear Broken Motors: Signature-Guided Data Augmentation for Induction-Motor Diagnostics}

\author[aff1]{Saraa Ali\fnref{fn1}}

\author[aff1]{Aleksandr Khizhik\fnref{fn2}}

\author[aff1]{Stepan Svirin\fnref{fn3}}

\author[aff1]{Artem Ryzhikov\fnref{fn4}}

\author[aff1]{Denis Derkach\fnref{fn5}}

\address[aff1]{National Research University Higher School of Economics, Moscow, Russia}

\fntext[fn1]{\texttt{thraaali@hse.ru}; ORCID: 0000-0002-1154-1262}

\fntext[fn2]{\texttt{akhizhik@hse.ru}; ORCID: 0009-0007-7695-9348}

\fntext[fn3]{\texttt{stequoy@gmail.com}; ORCID: 0009-0008-9339-3515}

\fntext[fn4]{\texttt{aryzhikov@hse.ru}; ORCID: 0000-0002-3543-0313}

\fntext[fn5]{\texttt{dderkach@hse.ru}; ORCID: 0000-0001-5871-0628}

\begin{abstract}
    The application of machine learning (ML) algorithms in the intelligent diagnosis of three-phase engines has the potential to significantly enhance diagnostic performance and accuracy. Traditional methods largely rely on signature analysis, which, despite being a standard practice, can benefit from the integration of advanced ML techniques. In our study, we innovate by combining ML algorithms with a novel unsupervised anomaly generation methodology that takes into account the engine physics model. We propose Signature-Guided Data Augmentation (SGDA), an unsupervised framework that synthesizes physically plausible faults directly in the frequency domain of healthy current signals. Guided by Motor Current Signature Analysis, SGDA creates diverse and realistic anomalies without resorting to computationally intensive simulations.
    This hybrid approach leverages the strengths of both supervised ML and unsupervised signature analysis, achieving superior diagnostic accuracy and reliability along with wide industrial application. The findings highlight the potential of our approach to contribute significantly to the field of engine diagnostics, offering a robust and efficient solution for real-world applications.
\end{abstract}


\end{frontmatter}



\section{Introduction}

The reliability and efficiency of three-phase engines is critical for numerous industrial applications, which makes their accurate and timely diagnosis essential. Traditional diagnostic methods for these engines predominantly rely on signature analysis, a technique that examines the engine's operational patterns to detect anomalies~\cite{book_motor}. Although signature analysis has become a de facto standard due to its effectiveness, it has some substantial limitations, and the growing complexity of modern engines and the vast amounts of data they generate require more advanced and precise diagnostic frameworks~\cite{KHANJANI2021108622}.

At the same time, machine learning (ML) and artificial intelligence (AI) have emerged as essential tools integrated into various aspects of modern industry and research, ranging from recommendation algorithms~\cite{roySystematicReviewResearch2022} to healthcare~\cite{s23094178}. Although the potential for advancement in industrial applications is immense, many real-world ML implementations are hampered by the scarcity or complete absence of fault-labeled datasets, as companies rarely allow destructive testing on expensive equipment \cite{9285283}. This situation forces researchers to choose between two extremes: purely unsupervised methods that handle unlabeled (healthy-only) data but often fail to capture subtle or incipient defects, or purely supervised methods that assume large labeled databases of real faults but are virtually impossible to collect.

To bridge this gap, we introduce \textbf{Signature-Guided Data Augmentation (SGDA)}, which combines unsupervised signature analysis principles with a supervised learning pipeline by injecting physically guided anomalies into normal data. This approach goes beyond naive or random anomaly injection by leveraging known engine parameters and the physics of induction motors to place synthetic ``peaks'' or other anomaly indicators at precise fault-relevant frequencies. In this way, SGDA produces training data that much more realistically mimic defects without physically damaging the motors. The result is an effective data augmentation strategy that preserves the nuanced spectral signatures relevant to early-stage or subtle fault detection. It thus surpasses existing purely unsupervised or purely supervised techniques by offering high accuracy in data-scarce conditions.

In contrast to alternative semi-supervised or random data augmentation approaches that often insert simplistic ``glitches,'' SGDA specifically injects frequency domain peaks tied to actual defect mechanisms. This deliberate physics-based sampling helps the model learn to distinguish incipient anomalies that might otherwise blend with normal operational frequencies. Hence, even in scenarios with few or no real fault examples, SGDA fosters robust generalization across changing loads, mechanical noise, and operational complexities. Ultimately, this hybrid solution integrates supervised ML advantages (clear classification targets) with unsupervised methodology (training on abundant healthy data) while retaining physically correct and diversity-rich anomalies.

In our study, we investigate this novel SGDA technique in the context of three-phase induction motors, also known for their robustness and reliability~\cite{IMARC2023}. These motors power pumps, compressors, conveyors, and fans across numerous industries~\cite{Siddiqui2014}. Despite their rugged design, unexpected failures—particularly subtle ones—can lead to costly downtime or unsafe operating conditions. Our experiments show that SGDA-trained models significantly outperform existing approaches (e.g., purely unsupervised anomaly detection~\cite{jeong2023anomalybert,9053558} or classical SVM-based classification \cite{1714134}) on real and synthetic datasets, even when labeled fault data is extremely limited. This provides a practical, physics-informed diagnostic method to detect and classify faults.

Below, we formally define the fault-detection problem for three-phase induction motors, review relevant literature, describe the SGDA methodology in detail, and present our experimental results. We conclude by discussing the implications of SGDA for large-scale industrial deployment, highlighting its ability to handle data scarcity while detecting subtle or incipient anomalies with higher fidelity than methods relying solely on naive random data augmentation or purely unsupervised feature extraction.

\section{Background}
This section reviews state-of-the-art solutions related to the fault detection problem in induction motors, focusing on signature analysis and machine learning-based approaches.

\subsection{Three-phase engines}
Three-phase engines\cite{aits2018em2, vssut2015em2}, also known as three-phase induction motors, are a fundamental component in various industrial applications due to their robustness, efficiency, and reliability. These motors operate on a three-phase power supply, which generates a rotating magnetic field within the stator. The stator is the stationary part of the motor and contains windings connected to the three-phase power supply. When the three-phase current flows through the stator windings, it induces a rotating magnetic field. The rotor, which is the spinning part of the motor, is placed within the stator. It typically consists of a laminated iron core with conductors, usually in the form of aluminum or copper bars, embedded within it. These conductors are arranged to form a closed loop known as a squirrel-cage rotor.
 
The operation of three-phase engines is based on the principles of electromagnetic induction. As the rotating magnetic field produced by the stator sweeps past the rotor, it induces an electromotive force (EMF) in the rotor conductors. According to Faraday's Law of Electromagnetic Induction, this induced EMF generates currents within the rotor. These currents interact with the magnetic field of the stator, producing a torque that causes the rotor to turn. The speed of the rotor is slightly less than the speed of the rotating magnetic field, a difference known as a slip, which is essential for torque production.

Three-phase induction motors are favored in industrial settings for several reasons. They offer high efficiency, providing a superior power-to-weight ratio compared to other motor types. Their simple and rugged construction makes them highly durable and capable of withstanding harsh operating conditions. Moreover, they require relatively low maintenance since they lack brushes and commutators found in other types of motors, reducing the wear and tear typically associated with these components.

The versatility and reliability of three-phase induction motors make them suitable for a wide range of applications. They are commonly used in machinery such as pumps, compressors, conveyors, and fans, where consistent and reliable operations are critical. Given their widespread use, the ability to accurately diagnose and maintain these motors is of paramount importance to ensure operational efficiency and minimize downtime in industrial processes. Traditional diagnostic methods, primarily based on signature analysis, have been effective but are increasingly challenged by the complexity and volume of data generated by modern engines. This requires the development of more advanced diagnostic tools, incorporating machine learning and artificial intelligence to enhance the precision and reliability of fault detection in three-phase induction motors.

\subsection{Engine Defect Types}

Three-phase induction motors, despite their robustness, can suffer from various defects that impact performance and lifespan \cite{Albrecht1986}. Understanding these defects is the key for effective diagnostics and maintenance. Common defect types include:

\paragraph{Intercell Shortages}

Intercell shortages, which account for $42\%$ of overall faults \cite{singhInductionMachineDrive2003}, are the result of short circuits between the turns of the stator windings, typically due to insulation breakdown from thermal stress, electrical surges, or physical damage. This defect causes uneven current distribution, localized heating, increased losses, and reduced efficiency, potentially leading to catastrophic failure if unaddressed.

\paragraph{Rotor Cell Defect}

Rotor cell defects involve faults in the rotor bars or end rings, caused by manufacturing flaws, thermal stress, or mechanical fatigue, account for approximately $10\%$ of all faults encountered in induction motors \cite{1546063}. These defects, such as broken bars or cracked rings, disrupt the motor's electromagnetic balance, leading to increased vibration, reduced torque, and uneven rotor heating, accelerating wear on other components.

\paragraph{Bearing Defect}

Bearing defects, which arise from inadequate lubrication, contamination, misalignment, or material fatigue \cite{Harris2006}, account for almost half of all induction motor failures \cite{4158071}. Symptoms include increased noise, vibration, and friction, leading to overheating and potential bearing seizure. Bearing failures can cause rotor misalignment and stator-rotor contact, necessitating costly repairs or replacements.

\paragraph{Eccentricity of the Air Gap}

Eccentricity of the air gap refers to uneven spacing between the stator and rotor, caused by manufacturing inaccuracies, bearing wear, or rotor deformation \cite{cameronj.r.VibrationCurrentMonitoring1986}. It can be static (fixed) or dynamic (variable). This defect leads to unbalanced magnetic pull, increased vibration, noise, and uneven wear, resulting in performance degradation and potential failure over time.

\paragraph{Unknown Mechanical Defects}

Unknown mechanical defects encompass issues like misalignments, loose components, structural weaknesses, and other unidentified problems. These defects often cause abnormal vibrations, noises, or irregular behavior, requiring comprehensive analysis and advanced diagnostics to accurately identify and rectify.

Each of these defect types presents unique challenges for the diagnosis and maintenance of three-phase induction motors. Developing advanced diagnostic methods, such as those incorporating machine learning and AI, can significantly enhance the detection and classification of these defects, leading to improved motor reliability and longevity.


\subsection{Overview of Fault Detection Methods}
Over the past decade, research on induction motor fault detection has evolved from classical signal analysis techniques to advanced machine learning (ML) and artificial intelligence (AI) methods. More recently, physics-informed data augmentation strategies have emerged, leveraging domain knowledge and simulations to improve diagnostic accuracy even with limited real fault data.
\subsection{Traditional Fault Detection Methods}

Early fault diagnosis in induction motors relied on direct monitoring of electrical and mechanical signals using signal processing and heuristic rules. Motor Current Signature Analysis (MCSA) is a prime example of a traditional non-intrusive technique that has been widely used to detect faults. MCSA involves analyzing the frequency spectrum of the stator current to identify characteristic frequency components associated with specific faults (e.g. sideband frequencies for broken rotor bars) \cite{s25020471,s22239494}. MCSA and related spectral analysis methods can effectively detect certain faults and even estimate motor speed in sensorless setups \cite{s25020471}. Similarly, vibration analysis, acoustic noise monitoring, and thermal imaging have been used to catch mechanical imbalances, bearing defects, or insulation failures by recognizing their distinctive signal patterns. These classical approaches require substantial expert knowledge to interpret results; for example, an expert may need to discern subtle sidebands in a Fourier spectrum or threshold an increase in vibration harmonics. Moreover, their performance often degrades under varying operating conditions. In the case of MCSA, changes in load or speed can mask or shift fault signature frequencies, making detection less reliable \cite{s25020471}. Techniques like basic FFT-based current analysis may fail for non-stationary operating regimes (e.g. load fluctuations) or incipient faults with very weak signatures \cite{en15238938}. Advanced time-frequency methods (wavelets, Wigner–Ville distributions, Hilbert transforms) have been introduced to handle non-stationary signals, but they add computational complexity \cite{en15238938}. This complexity can be problematic for online monitoring on low-cost hardware, partly explaining why many sophisticated signal processing techniques see limited deployment in industry \cite{en15238938}. In summary, traditional fault detection methods provide a foundation and are still used (MCSA is reported as one of the most effective classical methods \cite{s22239494}), but they suffer from key limitations: they are often labor-intensive, sensitive to operating condition variability, and limited in generalization (each method may be tuned to specific fault types or motors). These shortcomings set the stage for data-driven techniques that can automate and generalize fault detection.

Despite its proven applicability, a signature analysis has notable limitations. Certain fault signatures can be masked by operating conditions or other signal components. For instance, as shown above, fault-related frequencies may be too close to the primary frequency or hidden when the motor is lightly loaded, making detection challenging \cite{en15228569}.

\subsection{Machine Learning approaches for Fault Detection}

To overcome the limitations of purely analytical techniques, researchers have increasingly turned to machine learning to perform fault detection in induction motors. Early data-driven approaches combined signal feature extraction with classifiers: for example, computing statistical features or frequency components from motor signals and feeding them into Support Vector Machines (SVMs), artificial neural networks (ANNs), or $k$-Nearest Neighbor algorithms \cite{s22239494}. Such methods showed promising performance in identifying fault patterns, especially as computational power grew \cite{s22239494}. SVMs were among the first ML methods applied, valued for their low computation cost, though they could struggle in complex multi-fault scenarios \cite{s25020471}. These classical ML models still relied on human-crafted features and could be limited by the quality of those features.

In the last years, deep learning techniques have become prevalent for induction motor fault diagnosis, often bypassing manual feature engineering. Convolutional Neural Networks (CNNs)\cite{9451544} and Recurrent Neural Networks (RNNs) can learn directly from raw or minimally processed sensor data. For instance, researchers have applied 1D CNNs to raw current or vibration time-series to automatically learn fault-indicative patterns \cite{s25020471}. Others have transformed time-domain signals into 2D time-frequency images (spectrograms, wavelet scalograms) and used 2D CNNs for classification \cite{s25020471}. This \textit{signal-to-image} approach leverages CNNs’ strength in image recognition to capture subtle spectral features that might be missed by purely temporal models \cite{s25020471}. In one recent study, current signals from various conditions (healthy and multiple fault types) were converted to Fourier spectrum images and fed to a deep CNN (VGG-19 architecture), achieving up to 98–100\% classification accuracy for several fault classes \cite{s25020471}. The ability of deep networks to capture complex, nonlinear feature combinations has generally led to higher detection accuracy than earlier methods, as well as the ability to handle multiple fault categories simultaneously. For example, a CNN trained on high-resolution vibration spectra outperformed traditional handcrafted feature methods by more than 7\% precision in distinguishing normal vs. misalignment, unbalance, and bearing faults \cite{math9182336}. Recurrent models have also been explored for sequential data: Long Short-Term Memory (LSTM) \cite{graves2012long} networks improve on vanilla RNNs by retaining long-term temporal dependencies and have successfully detected faults like broken rotor bars from transient current signals \cite{s25020471}. More recently, Transformer-based models with attention mechanisms \cite{NIPS2017_3f5ee243} have been applied to motor current time-series, showing an ability to learn long-range relationships and identify fault signatures in scenarios with complex load variations \cite{FU2024381, s25020471}.

Despite these successes, purely data-driven models face certain challenges. Supervised deep learning models require large labeled datasets spanning all relevant operating modes and fault conditions to generalize well \cite{en15238938}. In practice, obtaining extensive fault data is challenging – motors do not fail often under controlled conditions, while intentional faults inducing (like breaking rotor bars or degrading insulation) for data collection is costly and sometimes impractical. This data scarcity often forces labs and industry to train models on limited fault examples, risking overfitting. Furthermore, models trained on one machine may not generalize to others without retraining, as different motor designs or inverter settings alter the signal patterns (the \textbf{domain shift} problem) \cite{en15238938}. In real industrial settings, a fault diagnosis system must handle varying loads, speeds, and even multiple simultaneous incipient faults – conditions under which an AI model might misclassify if those patterns were not well-represented in training data. These issues have catalyzed interest in \textit{physics-informed and augmented data approaches} to bolster machine learning models, marrying the strengths of domain knowledge with data-driven learning.

\subsection{Physics-Informed Data Augmentation}

Physics-informed data augmentation refers to techniques that incorporate knowledge of the motor’s physical behavior or use physics-based simulations to improve the training of diagnostic models. The core idea is to overcome limited or biased datasets by generating or selecting data in a way that honors the real fault physics, thereby improving model accuracy and robustness. Several strategies have emerged in recent years:
\begin{itemize}
    \item \textbf{Frequency-Domain Feature Injection}: A straightforward way to inform an AI model of motor physics is to feed it input representations that emphasize known fault characteristics. For example, Boushaba et al. approach uses Motor Current Signature Analysis preprocessing combined with CNNs for broken rotor bar detection~\cite{s22239494}. By transforming raw current signals to the frequency domain and supplying the CNN with spectra, the model is essentially forced to learn from physically meaningful fault harmonics (such as sideband frequencies due to a broken bar). This design yielded a physics-informed CNN that achieved 100\% detection of broken bars regardless of motor load or speed variations by locking onto robust fault indicators. Recent studies further enhance this strategy by introducing frequency-domain augmentation methods. Gwak et al. developed a Random Spectral Scaling technique that injects realistic spectral variability by randomly scaling frequency components—simulating shifts due to changing operating conditions—and then reconstructing the time-series signal via an inverse FFT while preserving conjugate symmetry~\cite{10003718}. Meanwhile, Huang et al. introduced a notch-filter augmentation that deliberately removes parts of the spectrum (e.g., the fundamental frequency), prompting the network to focus on subtle fault-related harmonics~\cite{en17163956}. These methods underscore that physics-informed frequency-domain feature engineering not only improves generalization across diverse operating conditions but also emphasizes the importance of realistic augmentation tailored to known fault signatures.
    \item \textbf{Synthetic Data Generation via Simulation}: When real fault data are scarce, a powerful approach is to generate virtual fault data using physics-based models and digital twins. High-fidelity simulation tools—such as finite element analysis (FEA) and multi-physics simulators—can emulate the electrical, magnetic, and vibrational responses under various fault conditions. For example, Zhang et al. addressed the scarcity of turbine rotor fault examples by using FEA to simulate vibration signals for different fault scenarios. In their FEMATL framework, 1D displacement signals were converted into 2D time-frequency spectrograms and then used as inputs for a deep classifier with transfer learning on a pre-trained ResNet18, achieving high diagnostic accuracy with only a small amount of real data~\cite{e25030414}. Similarly, Gao et al. combined FEA with generative adversarial networks (GANs) to improve bearing fault detection. Their approach merged FEM-generated vibration signals with GAN-generated samples to produce a rich, augmented dataset that improved diagnostic accuracy by approximately 7\%~\cite{MA2021115234}. Digital twins have also proven valuable for data augmentation. Zheng et al. built a Unity3D-based digital twin of an industrial reducer capable of simulating various fault types under different speed conditions. By integrating historical fault data with their physics model, they generated a comprehensive training set that enabled a deep learning model with a ConvNet backbone to achieve 99.5\% fault recognition accuracy in real industrial tests~\cite{s24082575}. Additional studies support these findings. For instance, Pasqualotto et al.~\cite{9547305} demonstrated that augmenting limited real data with synthetic signals greatly enhances fault detection performance. Although simulation-based augmentation can dramatically reduce the need for extensive physical prototypes, its success hinges o,s24082575n the simulation fidelity. Incomplete models or discrepancies between simulated and real-world behavior may introduce bias.
    \item \textbf{Domain Transfer and Hybrid Models}: A related trend is combining simulations with machine learning in hybrid frameworks that transfer knowledge from the virtual to the real domain. For example, Xia et al. (2023) introduced a digital twin-enhanced semi-supervised diagnosis framework for induction motors under label-scarce conditions~\cite{Xia2023}. Their method uses a multi-physics motor digital twin to generate simulated data, applies phase-contrastive current dot patterns (PCCDP) to convert three-phase currents into images, and performs inter-space sample generation to bridge simulation and real data. Similarly, Zhang et al. leveraged transfer learning by pretraining a ConvNext CNN on a balanced dataset of simulated bearing vibrations before fine-tuning on limited real data, thereby enhancing fault detection accuracy~\cite{s23115334}. Additionally, Liang et al. proposed a sparsity-constrained GAN (SC-GAN) that synthesizes hybrid signals from both simulation and real measurements, enriching the training dataset and reducing misclassification~\cite{MA2021115234}. These studies underscore how physics-informed augmentation combined with domain adaptation can significantly cut down the need for extensive real-world data in industrial diagnostics.
    \item \textbf{Augmentation of Measured Data}: Not all augmentation is via heavy simulations; some strategies simply manipulate the available data in a physics-informed way to create new training examples. One example is shifting window segmentation for time-series currents~\cite{en17163956}. Fu and Asrari demonstrated that instead of training a deep model on fixed-length current signal segments starting at arbitrary phase angles, one can slide the window to generate many overlapping segments, effectively augmenting the dataset and exposing the model to many phase conditions. This prevents the network from becoming biased to a particular sampling alignment (a form of augmentation that respects the periodic nature of AC waveforms). They found that models trained with such shifting windows and random cropping of the signals generalized better to unseen operating conditions. Other common augmentations include adding realistic noise~\cite{jsan13050060}, scaling signal amplitudes, or injecting minor frequency shifts – all grounded in physical variations that might occur in practice (sensor noise, load changes, etc.).
\end{itemize}
Despite their utility, existing physics-informed augmentation strategies share several notable limitations. Most approaches,whether based on frequency-domain manipulation, simulation, or domain transfer,are inherently constrained by the availability of real fault data or high-fidelity motor models. Frequency-domain feature injection techniques rely on prior fault instances to guide augmentation. Simulation-based methods, while powerful, are computationally expensive and require detailed knowledge of motor geometry and dynamics, making them difficult to scale or generalize across different motor types. Furthermore, these methods often perform static, offline augmentation, limiting their ability to introduce dynamic variation during training. Critically, none of these approaches eliminate the dependency on labeled fault samples—a key bottleneck in practical industrial applications. 

To address these challenges, we propose the \textit{Signature-Guided Data Augmentation (SGDA)} framework, which uniquely combines domain knowledge from Motor Current Signature Analysis with an online augmentation strategy that operates entirely on healthy current signals. SGDA introduces synthetic, physically plausible fault signatures directly into the frequency domain during training, enabling the creation of robust diagnostic models without requiring any real faulty data. This capability not only reduces reliance on costly fault acquisition but also allows for flexible deployment across diverse motor configurations.

\section{Signature-Guided Data Augmentation (SGDA)}
 
Despite the advancements in both classical signal-based methods like Motor Current Signature Analysis (MCSA) and modern machine learning techniques, fault detection in three-phase induction motors still faces a critical bottleneck: the scarcity and variability of real-world fault data. Traditional MCSA can miss subtle or masked anomalies under varying operating conditions, while purely data-driven approaches typically require large, diverse, and labeled datasets — which are costly or impractical to obtain. Although recent physics-informed augmentation strategies offer partial relief by simulating known fault behaviors, many rely on computationally expensive simulations or lack generalization to unseen conditions.

To address these limitations, we propose a novel hybrid framework called Signature-Guided Data Augmentation (SGDA). SGDA integrates the domain knowledge of MCSA with the generalization strength of deep learning models. SGDA operates directly in the frequency domain, injecting synthetic but physically grounded fault features—guided by known spectral characteristics—into normal current signals. These augmentations create diverse, \textit{realistic} fault scenarios without requiring real fault data.

\begin{figure}[H]
    \centering
    \includegraphics[width=0.7\linewidth]{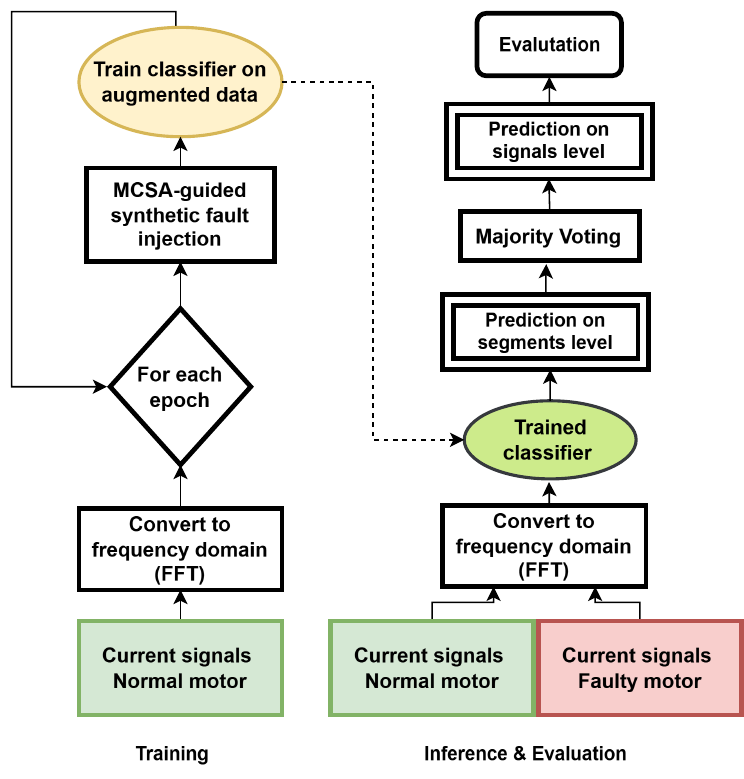}
    \caption{Overview of the SGDA framework. During training (left), only current signals from healthy motors are used; synthetic fault signatures are injected dynamically per epoch using MCSA-guided principles in the frequency domain. The model is trained on this augmented data. During inference (right), the trained classifier evaluates raw current signals—healthy or faulty—on a segment basis. Final predictions are aggregated using majority voting for robust signal-level fault classification.}
    \label{fig:sgda-framework}
\end{figure}

Figure~\ref{fig:sgda-framework} illustrates the SGDA framework. During training (left), only current signals from healthy motors are required. These signals are first transformed via Fast Fourier Transform (FFT). Then, during each training epoch, MCSA-guided synthetic fault injection is applied. A deep neural network classifier is trained on these augmented samples. The on-the-fly augmentation ensures continual variation in the training data, enhancing the model's ability to learn robust representations of both normal and fault conditions.

During inference (right), the trained classifier is applied to both normal and faulty motor signals. Incoming signals are segmented into 1-second windows, each processed individually. A final prediction is determined by aggregating segment-level outputs using a majority voting scheme—ensuring resilience to noise and localized signal anomalies.

\subsection{MCSA-guided synthetic fault injection}

Motor Current Signature Analysis (MCSA) plays a foundational role in the proposed Signature-Guided Data Augmentation (SGDA) framework by guiding where synthetic anomalies should be placed in the frequency spectrum. Unlike generic data augmentation methods, SGDA leverages domain-specific knowledge derived from MCSA to identify frequency regions associated with different fault types. These regions are typically determined using motor design parameters such as slip, pole count, and supply frequency~\cite{s22239494}. The resulting frequency bands, specific to each fault type, serve as targets for anomaly injection during training. This ensures that augmented data remains consistent with known electromechanical behaviors, enhancing both realism and diagnostic value.

\begin{figure}[h!]
    \centering
    \begin{subfigure}[b]{0.5\textwidth}
        \centering
        \includegraphics[width=\textwidth]{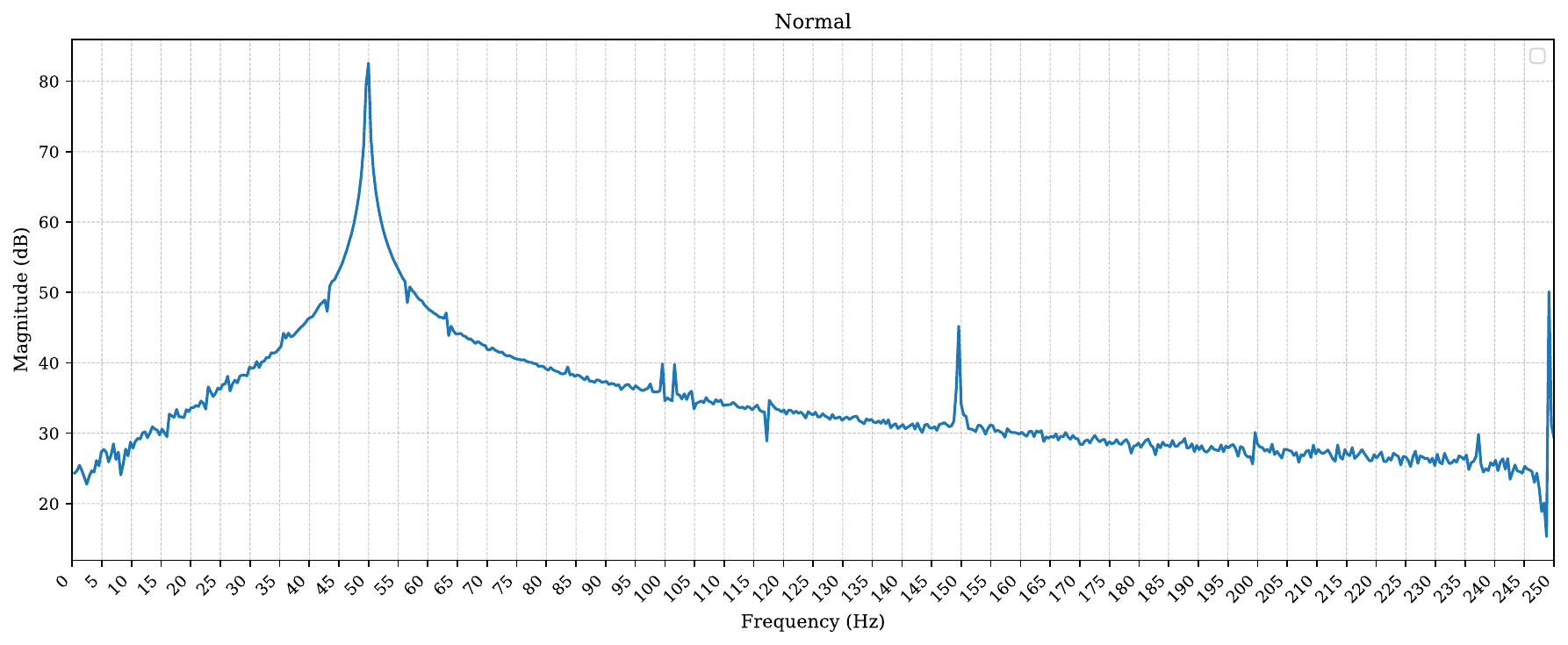}
        \caption{Normal motor condition}
    \end{subfigure}
    
    \vspace{0.5em}
    \begin{subfigure}[b]{0.49\textwidth}
        \centering
        \includegraphics[width=\textwidth]{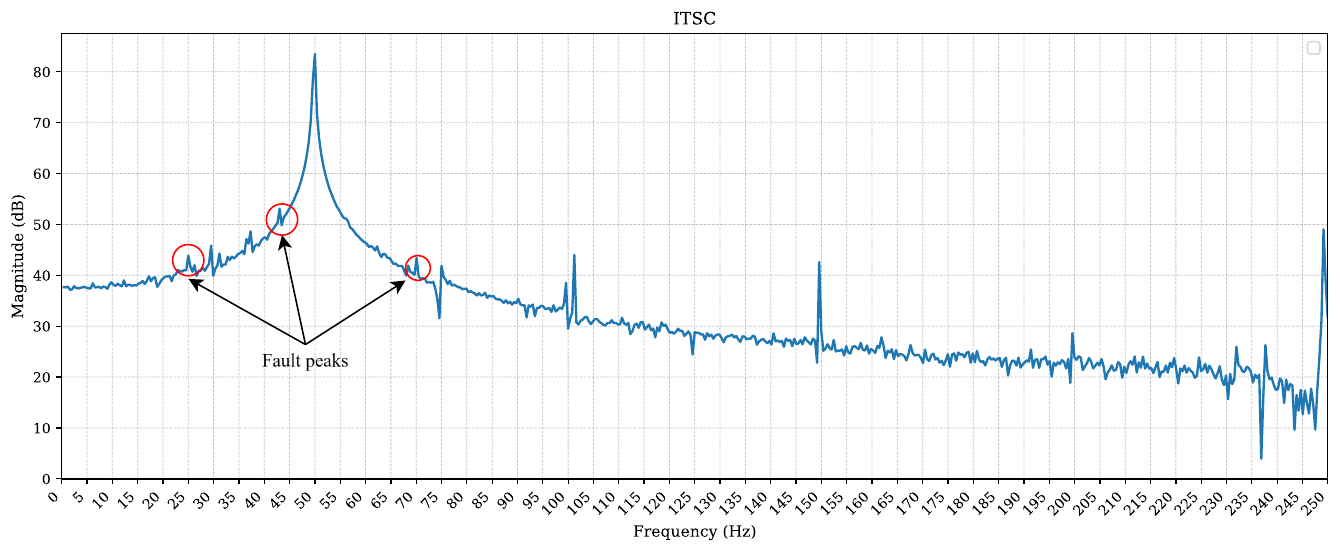}
        \caption{Inter-turn short circuit (ITSC)}
    \end{subfigure}
    \hfill
    \begin{subfigure}[b]{0.49\textwidth}
        \centering
        \includegraphics[width=\textwidth]{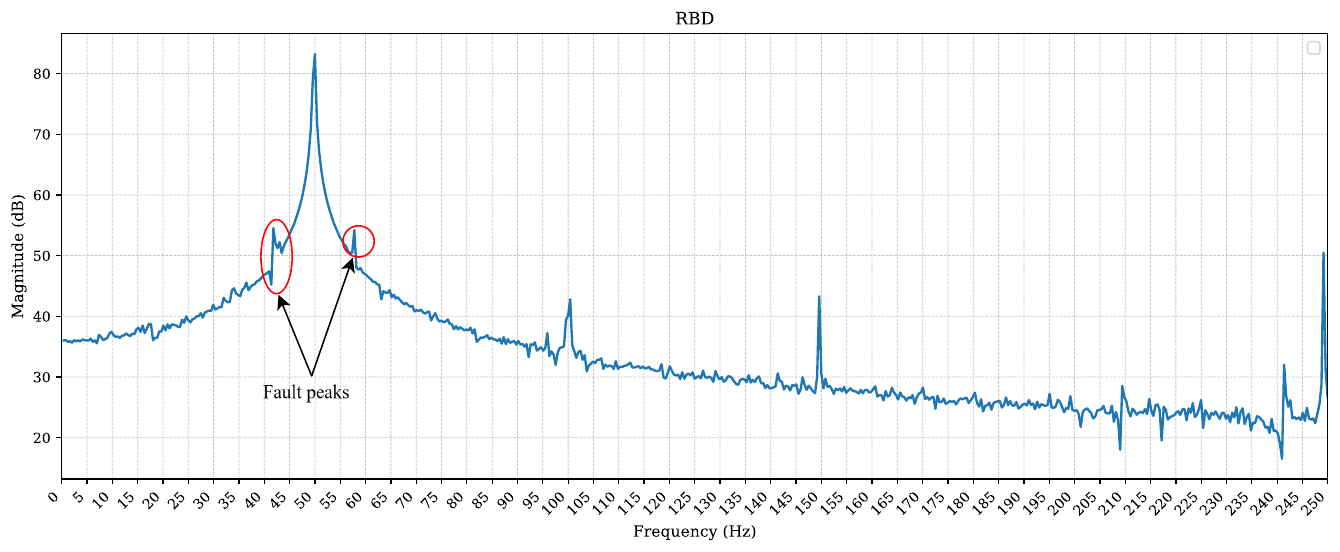}
        \caption{Rotor bar defect (RBD)}
    \end{subfigure}

    \caption{FFT spectra of motor current signals. (a) Normal condition with only base harmonics. (b) ITSC fault produces lower-frequency sidebands. (c) RBD fault exhibits distinct spectral asymmetries. Red circles mark MCSA-indicated fault zones used in SGDA augmentation.}
    \label{fig:mcsa-guided-peaks}
\end{figure}

As illustrated in Fig.~\ref{fig:mcsa-guided-peaks}, three FFT spectra are presented: (a) a normal operating motor, (b) a motor with an inter-turn short circuit (ITSC), and (c) one with a rotor bar defect (RBD). The circled regions in (b) and (c) indicate fault-specific spectral anomalies identified by MCSA. These fault signatures manifest as sidebands around key harmonics, and they serve as the injection templates within SGDA. By dynamically injecting Gaussian-shaped peaks into these regions (±~shift to account for variability), SGDA introduces realistic fault patterns into otherwise healthy signals.


\paragraph{Synthetic Fault Injection Guided by MCSA}
To generate realistic training samples that reflect typical motor faults, our SGDA framework leverages Motor Current Signature Analysis (MCSA) as a physics-based foundation for anomaly injection. Fault-related frequencies derived from MCSA serve as anchors where synthetic anomalies—specifically Gaussian-shaped peaks—are introduced into the FFT spectrum of healthy motor signals. Each injected peak captures key spectral characteristics associated with a specific fault type, such as harmonic patterns or sidebands near fundamental components.

This injection process is carefully designed to mirror the physical behavior of real faults, which typically manifest as smooth spectral modulations rather than sharp discontinuities. By introducing controlled variation in peak shape and intensity, SGDA not only captures known fault behaviors but also emulates subtle variations that may arise due to incipient conditions, operational noise, or parameter uncertainties. For more details see Section~\ref{sec:methodology_peak_injection}.
 
\begin{figure}[h!]
    \centering
    \includegraphics[width=\linewidth]{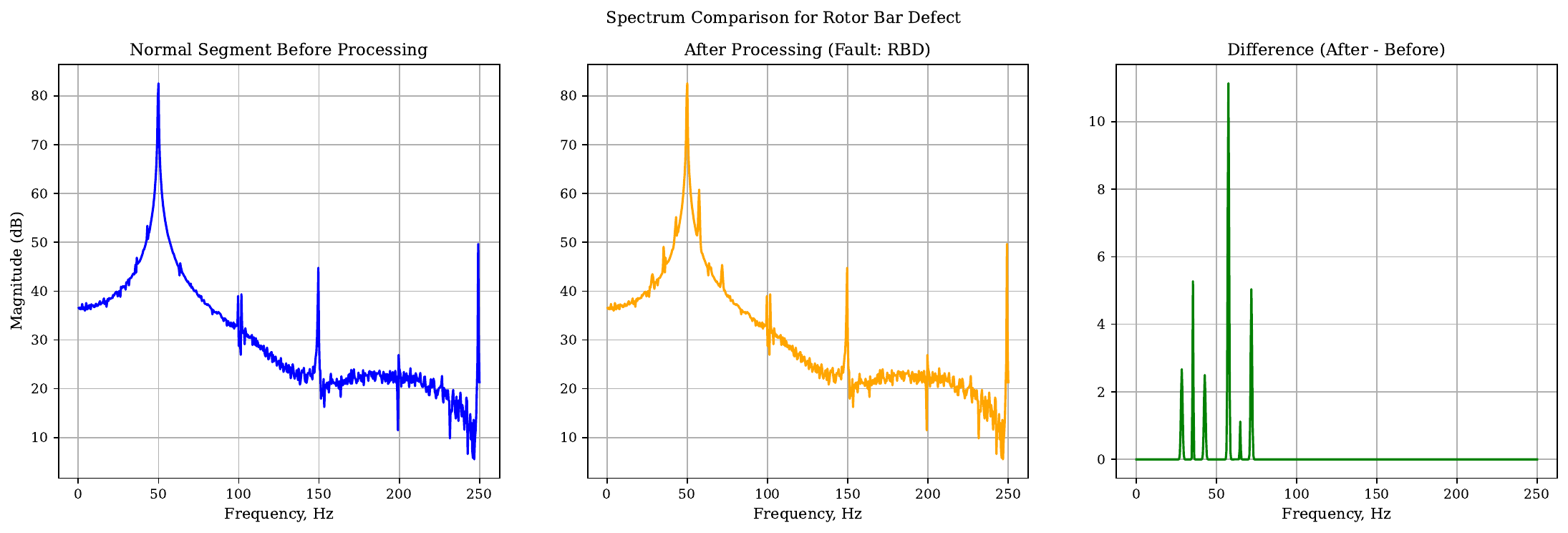}
    \caption{Illustration of SGDA augmentation. Left: original FFT spectrum of a healthy signal. Middle: augmented signal simulating a rotor bar fault. Right: difference plot showing localized injected peaks.}
    \label{fig:spectrum-augmentation-diff}
\end{figure}

As illustrated in Fig.~\ref{fig:spectrum-augmentation-diff}, this process produces localized spectral changes that reflect realistic fault characteristics. The figure shows a comparison between the original and augmented frequency spectra, along with the difference plot highlighting the injected anomalies.

\subsection{Deep Learning Integration}
\label{subsec:dl-integration}

As illustrated in Figure~\ref{fig:sgda-framework}, the SGDA framework is designed to be classifier-agnostic, enabling seamless integration with any supervised learning model. To evaluate this flexibility, we benchmarked several models including Logistic Regression, SVM, CatBoost, CNN, and ResNet18. For detailed comparative results, refer to Section~\ref{subsec:classifier-experiments}.

Based on its superior performance on binary and multiclass tasks, we selected a 1D ResNet-18 architecture for all subsequent experiments. Our implementation comprises four residual blocks, each consisting of two convolutional layers with pre-activation (Batch Normalization followed by ReLU).

Two ResNet models are independently trained: one for binary classification (normal vs. anomalous) and another for multiclass classification (normal, ITSC, and RBD). Both are optimized using the Adam optimizer with an initial learning rate of \(1 \times 10^{-3}\).
Learning rate scheduling is managed via \texttt{ReduceLROnPlateau}, and we use binary cross-entropy and categorical cross-entropy losses for binary and multiclass cases, respectively.

\textit{Additional ablation studies exploring dropout, attention mask, and other training configurations are provided in the \hyperref[appendix:experiments]{Appendix~B}.}

\subsection{Majority Voting – Inference and Classification Strategy}
\label{subsec:majority-voting-inference}

In practical settings, raw current signals often exhibit non-stationary behavior due to transient load changes, electromagnetic noise, or sensor drift. Consequently, not every short segment of a signal may contain sufficient fault-relevant information. To address this, our inference strategy is designed to aggregate predictions across multiple segments, enabling more stable and accurate signal-level decisions.
\paragraph{Segmentation and Prediction}
After training on SGDA-augmented data, inference begins by dividing each incoming current signal into 1-second segments. Each segment is transformed into the frequency domain via FFT and passed through the trained classifier—either binary (normal vs. fault) or multiclass (normal, ITSC, RBD).
Each 1-second segment yields an individual prediction, but since fault indicators can vary across a signal, per-segment predictions alone may be unreliable.
\paragraph{Signal-Level Inference via Majority Voting}
To infer the overall fault condition, we apply a majority voting strategy across all segment predictions of a signal. The final signal-level classification corresponds to the class that appears most frequently among the predicted segments. This mitigates the impact of transient misclassifications caused by noisy or ambiguous intervals, effectively filtering out spurious detections.
This decision-level smoothing is particularly beneficial in real-world diagnostics, where consistency over time is more valuable than single-frame precision. It aligns with the physical nature of motor faults, which generally manifest as sustained spectral patterns rather than isolated bursts.
\paragraph{Empirical Impact}
The benefits of this approach are empirically validated in Section~\ref{subsec:majority-voting-experiments}, where we show that majority voting consistently improves classification accuracy across varying phases and load conditions in both binary and multiclass settings.

\section{SGDA Methodology}

\subsection{Problem Formulation}
Let the engine be characterized by a parameter vector 
\[
\boldsymbol{\theta} \in \Theta \subset \mathbb{R}^d,
\]
where \(\Theta\) is assumed to be a subset of \(\mathbb{R}^d\). We introduce the engine state detection function
\[
f \colon \mathcal{X} \times \Theta \to \mathcal{Y},
\]
which assigns to every pair consisting of an observation and a corresponding parameter vector an engine state. Here, 
\begin{itemize}
    \item \(\mathcal{X}\) represents the set of all operational observations, encompassing measurements such as electrical current, vibration signals, and other pertinent characteristics,
    \item \(\mathcal{Y}\) denotes the set of all possible engine states. Under the assumption that only a single fault can occur at any given time, the set \(\mathcal{Y}\) is partitioned as
    \[
    \mathcal{Y} = \{\text{Normal}\} \cup \mathcal{T},
    \]
    where \(\mathcal{T}\) is the set of distinct fault types (for example, \(\{\text{Bearing Fault},\, \text{Inter-Turn Short Circuit},\, \dots\}\)).
\end{itemize}

\subsection{Motor Current Signature Analysis}
Motor Current Signature Analysis (MCSA) is a rigorous framework applied to engine fault diagnostics through the examination of motor current signals. Let \( \boldsymbol{\theta} \in \Theta \) denote the engine parameter vector, and let \( t \in \mathcal{T} \) represent a specific fault type. The MCSA methodology establishes an intrinsic correspondence between fault manifestations and their corresponding frequency signatures derived from the current spectrum.

To formalize this concept, we define the mapping
\[
\nu \colon \Theta \times \mathcal{T} \rightarrow \mathcal{P}(\mathbb{R}),
\]
which assigns to each ordered pair \( (\boldsymbol{\theta}, t) \) a subset of real numbers, representing the characteristic fault frequencies extracted from the motor current spectrum:
\[
\nu(\boldsymbol{\theta}, t) = \Bigl\{ \nu_1(\boldsymbol{\theta}, t),\, \nu_2(\boldsymbol{\theta}, t),\, \dots,\, \nu_{n(\boldsymbol{\theta}, t)}(\boldsymbol{\theta}, t) \Bigr\}.
\]
Here, \( n(\boldsymbol{\theta}, t) \) denotes the number of distinct fault frequencies associated with the parameter configuration \( \boldsymbol{\theta} \) and fault type \( t \). A detailed exposition of the methods utilized for the computation of these frequencies, along with parameter specifications, is provided in the \hyperref[appendix:mcsa]{Appendix~A}.

\subsection{Raw Signal Acquisition, Segmentation, and Fourier Transform}
\label{subsection:rsa}
The raw signal is modeled as a continuous, multidimensional time series 
\[
\mathbf{x}(t) \in \mathbb{R}^d,
\]
where \(d\) denotes the number of sensor channels and each component \(x_j(t)\) (for \(j = 1, 2, \dots, d\)) represents the measurement from the \(j\)th sensor at time \(t\).

Assuming the signal is sampled at discrete time instants \(t = nT_s\), where \(T_s > 0\) is the sampling period, the discrete-time representation is given by
\[
\mathbf{x}[n] = \bigl[x_1(nT_s),\;x_2(nT_s),\;\dots,\;x_d(nT_s)\bigr]^{\top}
\;\in\;\mathbb{R}^d, \quad n = 0,1,\dots,N-1.
\]
The discrete signal is further partitioned into overlapping segments via a sliding window. Let \(L\) denote the number of samples per segment and \(\Delta\) the step size (in samples) between successive segments. The \(i\)th segment is formally defined as
\[
X_i = \left[ \mathbf{x}[i\Delta],\, \mathbf{x}[i\Delta+1],\, \dots,\, \mathbf{x}[i\Delta+L-1] \right]^\top \in \mathbb{R}^{L \times d}, \quad i \in I,
\]
with the index set
\[
I = \left\{ 0, 1, \dots, \left\lfloor \frac{N - L}{\Delta} \right\rfloor \right\}.
\]

Subsequently, each segment is transformed into the frequency domain through the application of the Fast Fourier Transform (FFT) to each sensor channel independently. For the \(j\)th channel of segment \(i\), the FFT is computed as
\[
X_{i,j}^{\mathcal{F}} = \mathcal{F}\{ X_{i,j} \} \in \mathbb{C}^{N},
\]
where \(X_{i,j}\) denotes the \(j\)th column of \(X_i\). The FFT yields complex coefficients corresponding to the frequency bins
\[
\mathbf{f} = \{ f_k = k f_s / N : \; k = 0, 1, \dots, N-1 \},
\]
with \(f_s = \frac{1}{T_s}\) representing the sampling frequency and \(N\) the number of FFT points.

To improve numerical stability and accommodate a wide dynamic range, the magnitude spectrum is converted to a decibel scale:
\[
X_{i,j}^{\mathrm{dB}} = 20 \log_{10}\Bigl(|X_{i,j}^{\mathcal{F}}| + \epsilon\Bigr),
\]
where \(\epsilon > 0\) is a small constant introduced to avoid singularities. 

Finally, the decibel-scaled spectrum is normalized by min–max per channel:

\begin{itemize}
  \item \textbf{Individual segment–channel normalization:}
  \[
    \hat{X}_{i,j}
    = \frac{X_{i,j}^{\mathrm{dB}}
           - \min X_{i,j}^{\mathrm{dB}}}
           {\max X_{i,j}^{\mathrm{dB}}
           - \min X_{i,j}^{\mathrm{dB}}}.
  \]

  \item \textbf{Global per-channel normalization:}
  First compute channel-wise extrema over all segments,
  \[
    X_{\min}^{(j)}
    = \min_{i\in I} \,\bigl(\min X_{i,j}^{\mathrm{dB}}\bigr),
    \quad
    X_{\max}^{(j)}
    = \max_{i\in I} \,\bigl(\max X_{i,j}^{\mathrm{dB}}\bigr).
  \]
  Then
  \[
    \tilde{X}_{i,j}
    = \frac{X_{i,j}^{\mathrm{dB}} - X_{\min}^{(j)}}
           {X_{\max}^{(j)} - X_{\min}^{(j)}}.
  \]
\end{itemize}

Finally, we normalize the decibel spectrum by min–max scaling per channel and segment:
\[
\hat{X}_{i,j}
= \frac{X_{i,j}^{\mathrm{dB}} - m_{i,j}}
       {M_{i,j} - m_{i,j}},
\]
where the extrema \((m_{i,j},M_{i,j})\) are defined as:

\begin{itemize}
  \item \textbf{Individual segment–channel normalization:}
    \[
      m_{i,j} = \min X_{i,j}^{\mathrm{dB}}, 
      \quad
      M_{i,j} = \max X_{i,j}^{\mathrm{dB}}.
    \]

  \item \textbf{Global per-channel normalization (across all segments):}
    \[
      m_{i,j} 
      = \min_{i'\in I}\bigl(\min X_{i',j}^{\mathrm{dB}}\bigr), 
      \quad
      M_{i,j} 
      = \max_{i'\in I}\bigl(\max X_{i',j}^{\mathrm{dB}}\bigr).
    \]
\end{itemize}

Both approaches yield \(\hat{X}_{i,j}\in[0,1]\) yielding a normalized spectral representation suitable for further analysis.

\subsection{Synthetic Anomaly Injection and Dataset Augmentation}
\label{sec:methodology_peak_injection}
A cornerstone of the proposed SGDA methodology is the rigorous augmentation of the dataset via synthetic anomaly injection in the frequency domain. This augmentation is physically grounded in the MCSA framework, whereby the fault frequency mapping $\nu(\boldsymbol{\theta}, \cdot)$ provides a precise determination of the characteristic fault frequencies. In this exposition, we restrict our analysis to the case of a single-channel current signal (i.e., \(d=1\)).

Formally, we introduce the augmentation operator
\[
\mathcal{A} \colon \mathbb{R}^{L} \times \Theta \to \mathbb{R}^{L},
\]
which maps a normalized segment \(\hat{X}_i \in \mathbb{R}^{L}\) to its augmented counterpart \(\tilde{X}_i\) as follows:
\[
\tilde{X}_i(\boldsymbol{\theta}) = \mathcal{A}(\hat{X}_i, \boldsymbol{\theta}).
\]

Given a label \(y_i\), the augmented spectral representation is defined by
\[
\tilde{X}_i(\boldsymbol{\theta}, y_i)=
\begin{cases}
\hat{X}_i, & \text{if } y_i = \text{Normal},\\[1ex]
\hat{X}_i + \displaystyle\sum_{f^* \in \nu(\boldsymbol{\theta}, y_i)} \mathbf{P}\bigl(f^*; \psi, \epsilon_f\bigr), & \text{if } y_i \neq \text{Normal},
\end{cases}
\]
where \(y_i\) denotes the class label and the function 
\[
\mathbf{P}\bigl(f^*; \psi, \epsilon_f\bigr) = \Bigl[P\bigl(f_k - f^*; \psi, \epsilon_f\bigr)\Bigr]_{k=0}^{N-1}
\]
represents a vectorized Gaussian peak. The scalar function
\[
P(x; \psi, \epsilon_f) = \mathbb{I}\bigl\{|x| \leq \epsilon_f\bigr\}\, A \exp\!\left(-\frac{(x-\mu)^2}{2\sigma_g^2}\right)
\]
is defined with the parameter set \(\psi = \{A, \mu, \sigma_g\}\). The amplitude \(A\), modeling the intensity of the fault signature,  and standard deviation \(\sigma_g > 0\), representing the dispersion of the anomaly around target frequency $f^*$,  are modeled as random variables independently sampled from the uniform distributions
\[
A \sim \mathcal{U}(A_{\min}, A_{\max}), \quad \mu \sim \mathcal{U}(-\epsilon_f, \epsilon_f), \quad \sigma_g \sim \mathcal{U}(\sigma_{\min}, \sigma_{\max}),
\]

where:
    \begin{itemize}
        \item \( A_{\min} \), representing the average local spectral variability.
        \item \( A_{\max}\), the maximum magnitude observed in the FFT spectrum.
        \item \(\mu\) is the shift from  $f^*$, adding variability to the peak location.
        \item \(\sigma_{\min} \), \(\sigma_{\max} \), defining the peak's width range.
        \item \(\epsilon_f > 0\) defines the effective frequency window around each target frequency $f^*$, within which Gaussian peaks are injected. This window serves a dual purpose: it reflects uncertainties in motor parameters (e.g., slip, rotor slot count) and accounts for variations due to changing operational conditions. From a signal processing standpoint, \(\epsilon_f\) ensures that the augmentation remains localized, preventing synthetic peaks from distorting the entire spectrum. 
    \end{itemize}
    The amplitude is further multiplied by a randomly selected sign from a discrete uniform distribution \( \{-1, +1\} \), introducing polarity variation.

\subsection{Dataset Construction} \label{sec:dataset_construction}
The fully augmented dataset, generated from \(S\) independently acquired signals, is formally constructed as
\[
D^{\text{aug}}(\boldsymbol{\theta}) = \bigcup_{s=1}^{S} \bigl\{ \left( \tilde{X}_i^{(s)}(\boldsymbol{\theta}, y_i^{(s)}),\, y_i^{(s)} \right) \,\bigm|\, i \in I \bigr\},
\]
where, for each signal \(s\), \(\tilde{X}_i^{(s)}(\boldsymbol{\theta}, y_i^{(s)})\) denotes the augmented segment corresponding to the \(i\)th segment, and \(y_i^{(s)}\) is the associated label.



Algorithm~\ref{alg:data_augmentation} outlines the procedure for data augmentation given the preprocessed segments.

\begin{algorithm}[h!]
\caption{Per-Epoch Augmentation for Preprocessed Segments}
\label{alg:data_augmentation}
\begin{algorithmic}[1]
\Require Preprocessed segments $\hat{X}_i^{(s)}$ for $s = 1,\dots,S$;
        fault type set $\mathcal{T}$ and mapping $\nu(\boldsymbol{\theta},\cdot)$;
        Gaussian parameters $A_{\min},A_{\max},\sigma_{\min},\sigma_{\max}$;
        peak bandwidth $\epsilon_f$;
        integers $K,R\!\ge 0$;
        segment index set $I$
\Ensure Augmented dataset $D^{\mathrm{aug}}_{\mathrm{epoch}}(\boldsymbol{\theta})$
\State $D^{\mathrm{aug}}_{\mathrm{epoch}}(\boldsymbol{\theta})\gets\emptyset$
\For{$s = 1$ \textbf{to} $S$}
    \ForAll{$i \in I$}
        \For{$r = 1$ \textbf{to} $R+1$}
            \Comment{\emph{$R$ identical normal copies}}
            \State Append $\bigl(\hat{X}_i^{(s)},\text{Normal}\bigr)$
        \EndFor
        \For{$k = 1$ \textbf{to} $K$}
            \Comment{\emph{$K$ synthetic Gaussian-fault variants}}
            \State \textbf{Sample} fault $t \sim \text{Uniform}(\mathcal{T})$
            \State $\Delta X \gets \mathbf{0}$
            \ForAll{$f^\ast \in \nu(\boldsymbol{\theta},t)$}
                \State \textbf{Sample} $A \sim \mathcal{U}(A_{\min},A_{\max})$
                \State \textbf{Sample} $\mu \sim \mathcal{U}(-\epsilon_f,\epsilon_f)$
                \State \textbf{Sample} $\sigma_g \sim \mathcal{U}(\sigma_{\min},\sigma_{\max})$
                \State $\Delta X \gets \Delta X + \mathbf{P}(f^\ast;\psi,\epsilon_f)$
            \EndFor
            \State $\tilde{X} \gets \hat{X}_i^{(s)} + \Delta X$
            \State Append $\bigl(\tilde{X},t\bigr)$
        \EndFor
    \EndFor
\EndFor
\State \Return $D^{\mathrm{aug}}_{\mathrm{epoch}}(\boldsymbol{\theta})$
\end{algorithmic}
\end{algorithm}

\paragraph{Choice of replication parameters $R$ and $K$}
To obtain a perfectly balanced dataset at the level of every mini-epoch,
we replicate each \emph{normal} segment $R$ times and generate exactly
$K$ \emph{synthetic fault} variants.  For binary classification
(\(|\mathcal{T}| = 1\)) we set
\[
  K = 1 + R,
\]
so that the numbers of Normal and Fault samples are equal
for every segment.  For multiclass classification we require the
same count for every class in \(\mathcal{Y}\);
consequently
\[
  K = |\mathcal{T}|\,\bigl(1 + R\bigr),
\]
yielding \(1+R\) samples of each fault type and \(1+R\) normal samples.

\section{Dataset Description}
\label{sec:dataset-description}

We utilize two datasets collected from distinct three-phase induction motors to evaluate the proposed SGDA framework.

\paragraph{Motor A (Primary Dataset)}  
The first motor is a lower-power unit (AIR 80B4 IM1081), rated at 1.5~kW and operating at 1390~RPM. It is equipped with 6204 bearings and powered by a 380~V / 50~Hz source. Current signals were sampled at 4098~Hz. The Motor Current Signature Analysis (MCSA) parameters are: rotor slip \(s = 0.073\), 2 pole pairs, and a rotor frequency \(f_r = 23.17\)~Hz.

The dataset consists of 341 three-phase current signals, each labeled under one of three operating conditions: Normal, Inter-Turn Short Circuit (ITSC), and Rotor Bar Defect (RBD). Signals were recorded at six mechanical load levels: 0\%, 20\%, 40\%, 60\%, 80\%, and 100\%. Table~\ref{tab:dataset-overview} summarizes the distribution of samples per condition and load level.

The dataset for the second motor comprises 341 three-phase current signals collected from identical induction engines under three operating conditions: Normal, ITSC, RBD. Signals were recorded at six mechanical load levels (0\%, 20\%, 40\%, 60\%, 80\%, and 100\%). Table~\ref{tab:dataset-overview} summarizes the number of signals per condition and load.

\begin{table}[ht]
  \centering
  \begin{tabular}{lrrrrrrr}
    \toprule
    \textbf{Condition} & \textbf{Total} & \textbf{0\%} & \textbf{20\%} & \textbf{40\%} & \textbf{60\%} & \textbf{80\%} & \textbf{100\%} \\
    \midrule
    Normal & 128 & 21 & 23 & 20 & 22 & 21 & 21 \\
    ITSC   &  77 &  0 &  0 &  0 & 28 & 23 & 26 \\
    RBD    & 136 & 21 & 21 & 23 & 25 & 22 & 24 \\
    \midrule
    \textbf{Total} & \textbf{341} & 42 & 44 & 43 & 75 & 66 & 71 \\
    \bottomrule
  \end{tabular}
  \caption{Overview of current signals by condition and load level for Motor A.}
  \label{tab:dataset-overview}
\end{table}

For the binary classification task, all fault categories (ITSC, RBD) were grouped under a single “anomalous” class. For the multiclass task, each fault was assigned a unique label, allowing the classifier to learn fault-specific distinctions.

\paragraph{Motor B (Generalization Dataset)}  
The second motor is a high-power industrial-grade induction motor (RA180L4/2Y3), rated at 17~kW and operating at 1470~RPM. It is equipped with 6309/6310 bearings and operates at a supply frequency of 50~Hz. The sampling rate for data collection was 10,000~Hz. Key MCSA parameters include: rotor slip \(s = 0.028\), 34 rotor bars, 2 pole pairs, and a rotor frequency \(f_r = 23.33\)~Hz.

This dataset includes three single-phase current recordings—one per condition: Normal, ITSC, and RBD. Specific load levels are not provided. While less extensive, this dataset is used to evaluate model generalization across motor types. For detailed results on this experiment, see Section~\ref{subsec:other-motor-experiment}.


\section{Experiments and Results}

\subsection{Experiments: Evaluating Majority Voting in Binary Classification}
\label{subsec:majority-voting-experiments}

To assess the benefit of majority voting for signal-level binary classification, we conducted controlled experiments using the SGDA  with a ResNet-18 classifier trained on all load levels and phases.

\paragraph{Setup}
Both the training and testing datasets include signal segments from all three motor phases and cover the full range of load levels (0\% to 100\%).

\paragraph{Inference Strategy}
\begin{itemize}
    \item \textbf{Segment-Level:} Each segment (1-second window) was individually classified.
    \item \textbf{Signal-Level:} The final label for each signal was determined via majority voting across all segments.
\end{itemize}

\paragraph{Performance comparisons}

\begin{figure}[h!]
    \centering
    \begin{subfigure}[b]{0.49\textwidth}
    \includegraphics[width=\linewidth]{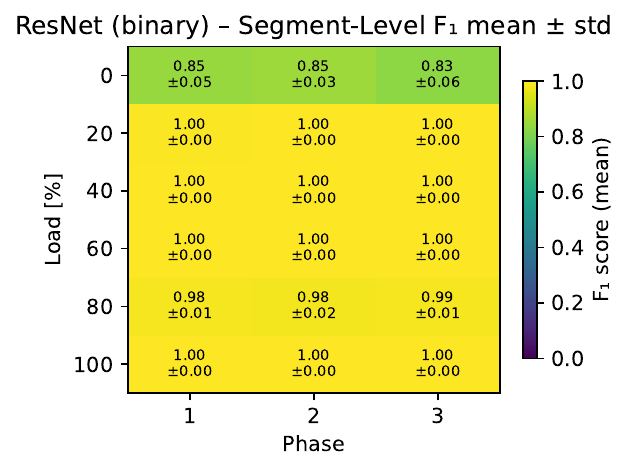}
    \caption{Segment-level $F_1$ (mean ± std)}
    \end{subfigure}
    \hfill
    \begin{subfigure}[b]{0.49\textwidth}
    \includegraphics[width=\linewidth]{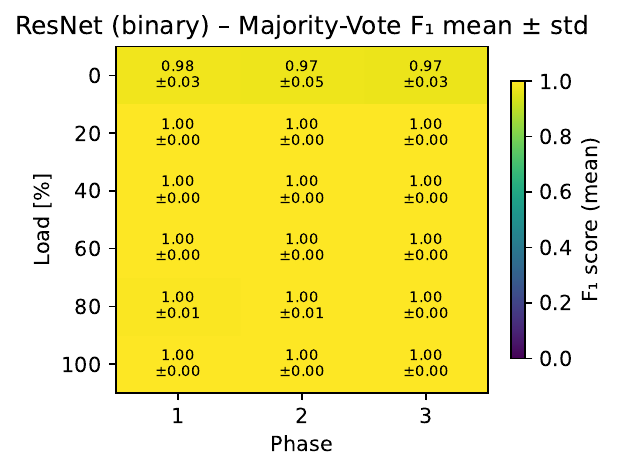}
    \caption{Signal-level (majority vote) $F_1$(mean ± std)}
    \end{subfigure}
    \caption{Binary classification performance with and without majority voting across load and phase. Majority voting clearly enhances classification robustness and consistency.}
    \label{fig:vote_compare}
\end{figure}

Figure~\ref{fig:vote_compare} presents the $F_1$ scores (mean ± standard deviation) across motor loads and input phases with and without majority voting. Segment-level performance shows a decline at 0\% load ($F_1$ between 0.83 and 0.85), highlighting vulnerability under minimal operational stress. Majority voting significantly boosts robustness, elevating scores to near-perfect values ($F_1$ between 0.97 and 1.00) and eliminating variability (\(\pm 0.00\)) in most cases.

These results validate majority voting as a good mechanism for increasing diagnostic stability and signal-level reliability, particularly under low-load and variable-phase scenarios.

\subsection{Experiments for Different Classifiers}
\label{subsec:classifier-experiments}

To assess the general applicability and flexibility of our SGDA framework, we evaluated multiple classifiers for both binary (normal vs. anomalous) and multiclass (normal, inter-turn short circuit (ITSC), rotor bar defect (RBD)) classification tasks. While SGDA can be integrated with any downstream classifier, this benchmarking provides insight into which architectures are best suited to capture spectral fault patterns. We tested a range of classifiers—including Logistic Regression, Support Vector Machines (SVM), CatBoost, a 1D CNN, and ResNet18—to explore trade-offs in performance, complexity, and generalization. Ultimately, although we adopt ResNet18 in our core pipeline due to its superior accuracy and robustness, the SGDA method is compatible with alternative models based on user preference or deployment constraints.

\paragraph{Setup}
All classifiers were trained and tested on the SGDA-augmented dataset, which included data from all phases and load levels (0\% to 100\%). Features were based on FFT-transformed current segments, consistent with the SGDA pipeline.

\paragraph{Results}
Table~\ref{tab:classifier-performance} presents a summary of performance metrics, including accuracy and macro F$_1$ scores. Traditional models like Logistic Regression and SVM exhibited limited learning capacity under the spectral feature domain. CatBoost showed improved performance due to its ability to model nonlinear interactions, especially in the binary task. However, its performance still lagged in multiclass settings, indicating difficulty in separating subtle spectral differences between fault types. Deep models, especially CNN and ResNet18, significantly outperformed others, with ResNet18 yielding the best results across both tasks.

\begin{table}[h!]
    \centering
    \begin{tabular}{lcccc}
        \toprule
        \textbf{Classifier} & \textbf{Task} & \textbf{Accuracy} & \textbf{Macro} $\mathbf{F_1}$ \\
        \midrule
        Logistic Regression & Binary    & 0.50  & 0.33 \\
        SVM (Linear)        & Binary    & 0.50  & 0.33 \\
        CatBoost            & Binary    & 0.86  & 0.86 \\
        CNN                 & Binary    & \textbf{0.98} & \textbf{0.98} \\
        ResNet18            & Binary    & \textbf{0.98} & \textbf{0.98} \\
        \midrule
        Logistic Regression & Multiclass & 0.31  & 0.31 \\
        SVM (Linear)        & Multiclass & 0.38  & 0.37 \\
        CatBoost            & Multiclass & 0.59  & 0.55 \\
        CNN                 & Multiclass & 0.74  & 0.72 \\
        ResNet18            & Multiclass & \textbf{0.80} & \textbf{0.79} \\
        \bottomrule
    \end{tabular}
    \caption{Classifier Performance Comparison on SGDA-Augmented Dataset}
    \label{tab:classifier-performance}
\end{table}

\begin{figure}[h!]
    \centering
    \includegraphics[width=0.48\linewidth]{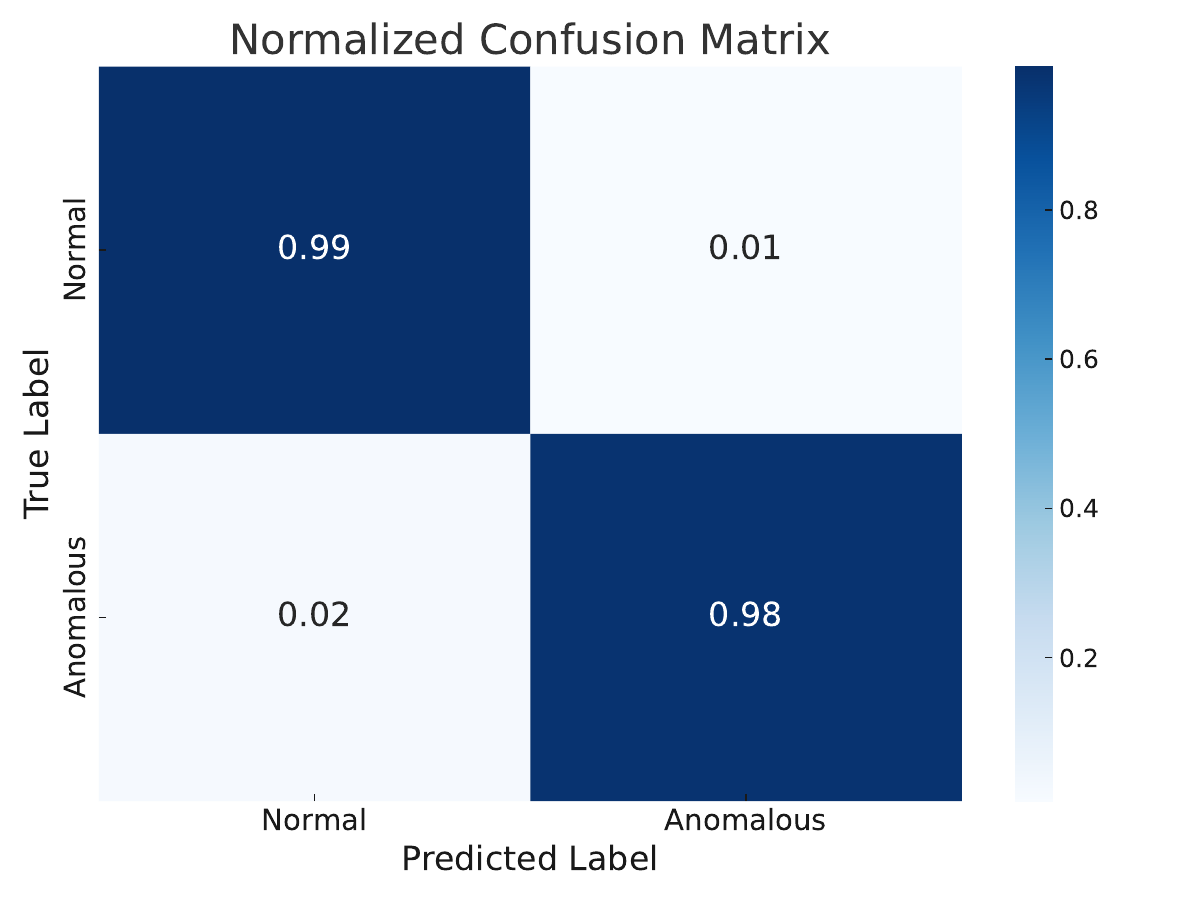}
    \includegraphics[width=0.48\linewidth]{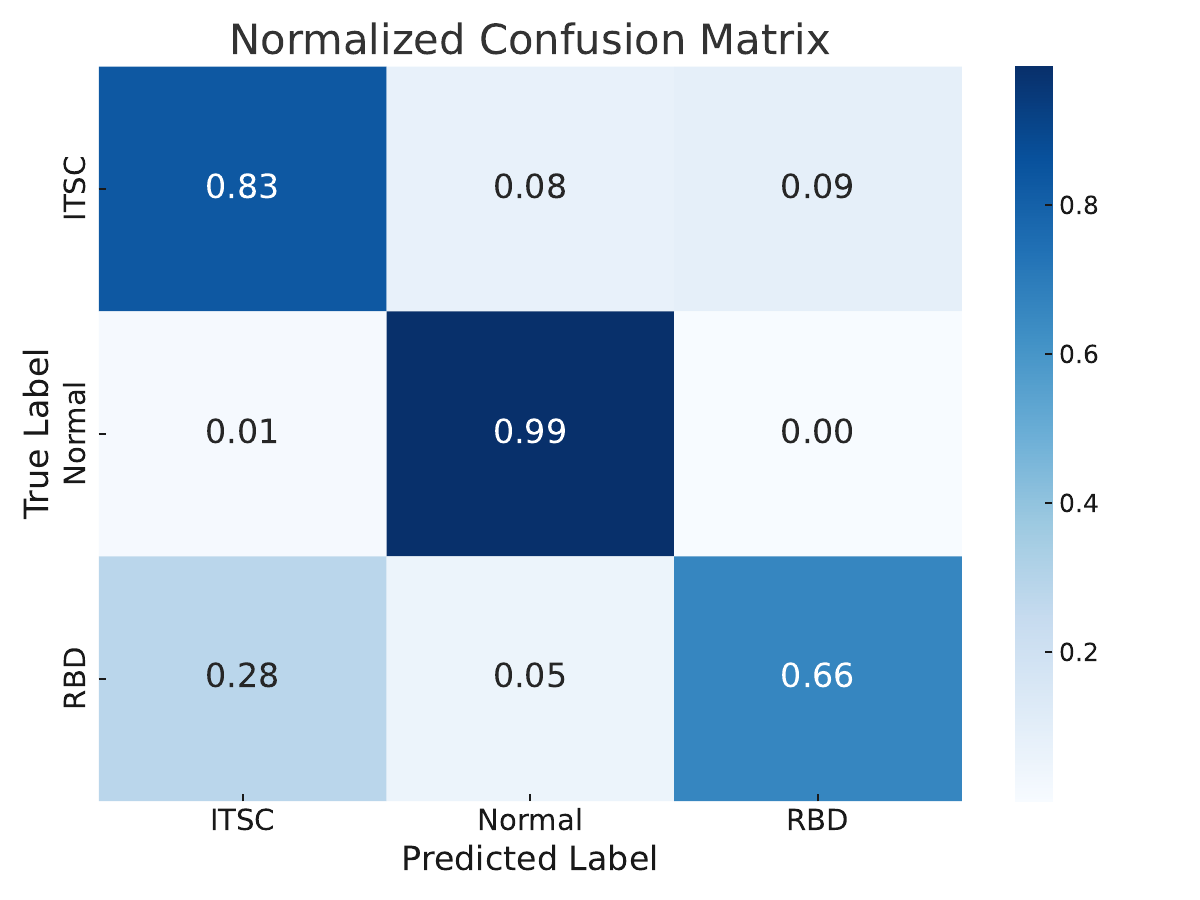}
    \caption{Confusion matrices for ResNet18: (left) binary classification, (right) multiclass classification. Results shown at the segment level.}
    \label{fig:confusion-matrices-resnet}
\end{figure}

\paragraph{Confusion Matrix Analysis}
To visually illustrate ResNet’s effectiveness, Fig.~\ref{fig:confusion-matrices-resnet} shows normalized confusion matrices for the ResNet18 classifier at the segment level. For binary classification, the model achieved 99\% precision on normal segments and 98\% on anomalous ones. In the multiclass setting, performance was strongest on the normal and ITSC classes, while RBD presented a greater challenge due to its subtler spectral signature overlap.

Given its strong performance across both tasks, ResNet18 was selected as the backbone classifier in subsequent SGDA evaluations. Its capacity to learn hierarchical spectral features makes it particularly suitable for FFT-based motor diagnostics.

\subsection{Generalization Across Load and Phase Conditions}
\label{subsec:load-phase-generalization}

To evaluate the generalization ability of the SGDA-trained classifier (ResNet18) across operational variations, we designed two experiments targeting both bbinary classification (normal vs. fault) and multiclass classification (normal, ITSC, RBD). 

\begin{figure}[h!]
    \centering
    \begin{subfigure}[b]{0.49\textwidth}
        \includegraphics[width=\textwidth]{figs/ResNet_binary_majority_mean_std_ALL.pdf}
        \caption{Binary – full training}
    \end{subfigure}
    \hfill
    \begin{subfigure}[b]{0.49\textwidth}
        \includegraphics[width=\textwidth]{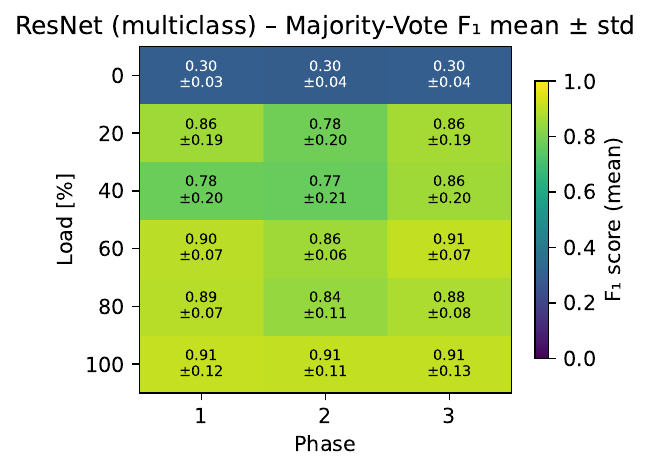}
        \caption{Multiclass – full training}
    \end{subfigure}
    \caption{\(F_1\) mean \(\pm\) std for full training across all phases and load levels.}
    \label{fig:load-phase-full}
\end{figure}

\paragraph{Experiment 1 – Training on All Loads and Phases}
In this setup, the ResNet model was trained using the full dataset, encompassing all three motor phases and load levels from 0\% to 100\%. Testing is likewise performed across the entire operational range. This represents an upper-bound configuration where the classifier has complete exposure to the variability present in the system.
Figure~\ref{fig:load-phase-full} presents \(F_1\) scores (mean \(\pm\) std) per phase and load level. In the binary setting, the model achieved near-perfect performance across all settings. Even at 0\% load, where classification is often more challenging due to weaker spectral fault signatures, \(F_1\) scores exceeded 0.97, with low variance, highlighting the robustness of the SGDA-trained model.

\begin{figure}[h!]
    \centering
    \includegraphics[width=0.45\linewidth]{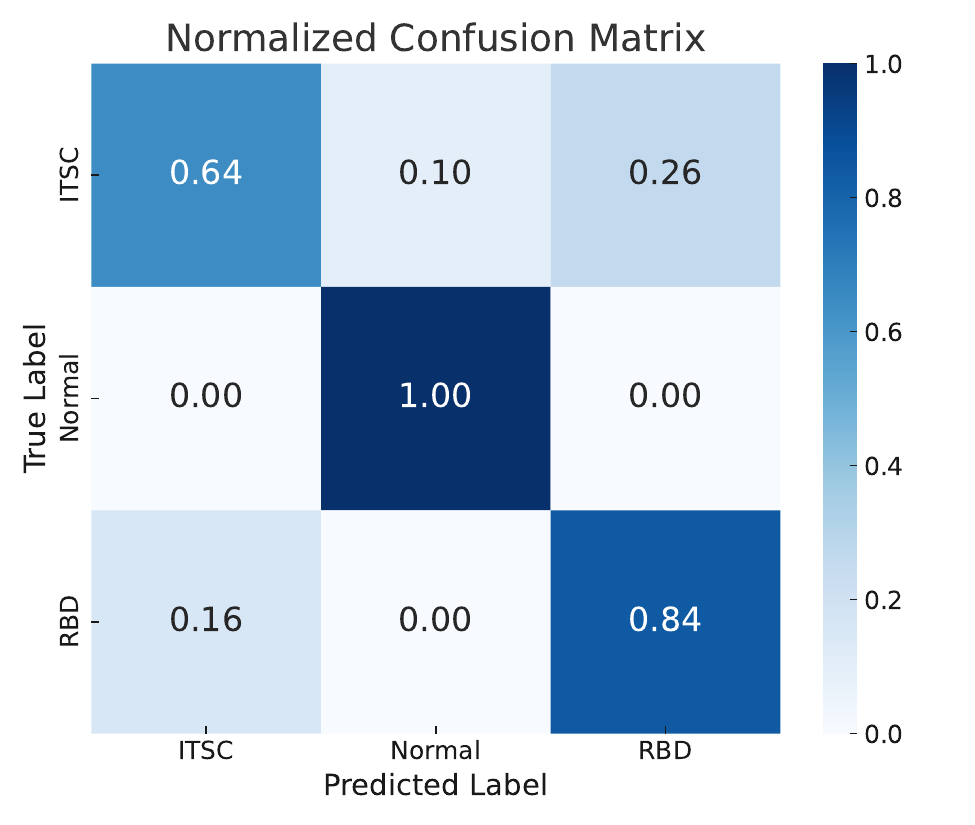}
    \caption{Signal-level normalized confusion matrix for the multiclass task under full training. Misclassification occurs between fault types (ITSC and RBD) but not with the normal class.}
    \label{fig:conf-matrix-multiclass-full}
\end{figure}

For the multiclass task, Performance was strongest at medium to high loads (\(\geq 60\%\)), where macro \(F_1\) scores consistently exceeded 0.88 across all phases with minimal variance. At lighter loads (20–40\%), performance became more variable, particularly on Phase 2, where standard deviations reached up to 0.21 and \(F_1\) dipped to 0.77. This is consistent with the attenuation of fault-related spectral features at lower torque, which makes fault type separation more difficult. At 0\% load, the model achieved an \(F_1\) score of approximately 0.30 across all phases, indicative of random guessing among three equally probable classes. However, as shown in the signal-level confusion matrix (Fig.~\ref{fig:conf-matrix-multiclass-full}), the model predominantly misclassified one fault type as another (e.g., RBD vs. ITSC), while maintaining perfect separation from the normal class. This finding confirms that even when uncertainty arises in fault-type discrimination, SGDA ensures high reliability in identifying whether a sample is faulty or not—critical for practical deployments.

\paragraph{Experiment 2 – Training on Phase 1 at 100\% Load Only}
Here, the classifier was trained using a highly constrained configuration: Phase 1 at 100\% load. Testing, however, was performed across all other loads and phases to evaluate the model’s generalization under minimal training diversity.

\begin{figure}[h!]
    \centering
    \begin{subfigure}[b]{0.49\textwidth}
        \includegraphics[width=\textwidth]{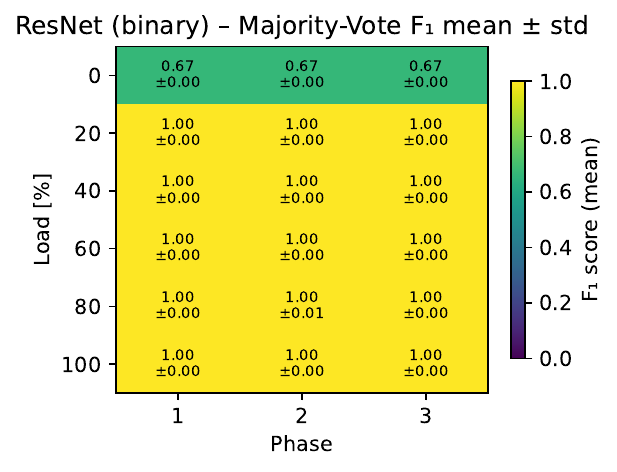}
        \caption{Binary – trained only on Phase 1 @ 100\%}
    \end{subfigure}
    \hfill
    \begin{subfigure}[b]{0.49\textwidth}
        \includegraphics[width=\textwidth]{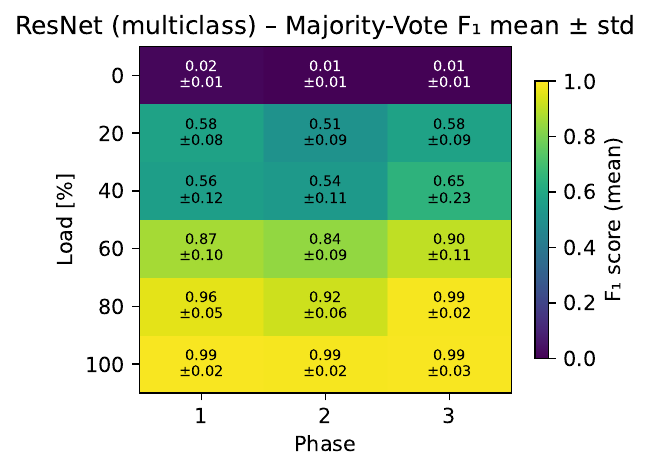}
        \caption{Multiclass – trained only on Phase 1 @ 100\%}
    \end{subfigure}
    \caption{\(F_1\) mean \(\pm\) std for limited training on one phase and one load.}
    \label{fig:load-phase-single}
\end{figure}

As shown in Fig.~\ref{fig:load-phase-single}, which presents macro \(F_1\) scores with standard deviation, binary classification generalized very well across all settings, with slight reductions at 0\% load (\(F_1 \approx 0.67\)). This robustness under binary labeling indicates that the model effectively learned to distinguish normal from faulty behavior, even from a narrow training subset.

The multiclass classifier exhibited greater sensitivity, with significant drops in performance under low loads, especially 0\% (\(F_1 \approx 0.02\)). Nevertheless, performance improved above 40\% load, reaffirming SGDA’s value even in resource-constrained training regimes. The model struggled to generalize under low-load conditions, however, at higher loads \(F_1\) scores exceeded 0.84 across all phases, peaking near 0.99 at full load. This suggests that while SGDA provides realistic spectral patterns, multiclass discrimination still benefits from diverse training conditions, especially at low torque.

\begin{figure}[h!]
    \centering
    \includegraphics[width=0.45\linewidth]{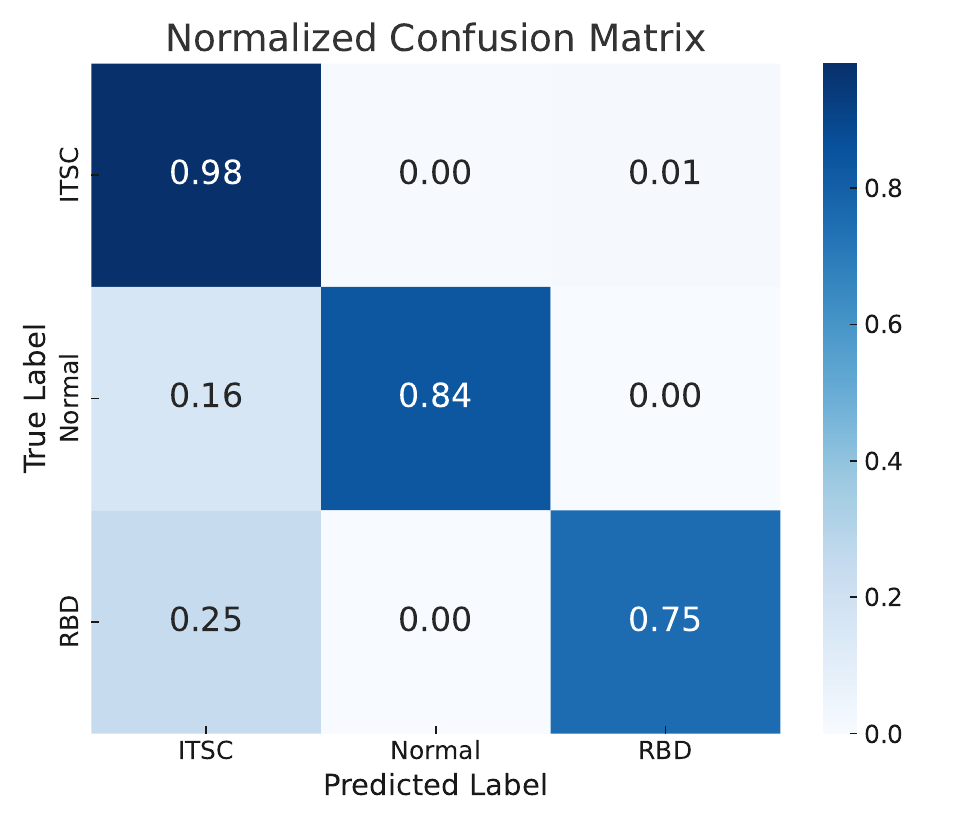}
    \caption{Signal-level normalized confusion matrix for the multiclass task trained only on Phase 1 @ 100\% load. Misclassifications are limited to fault types, not normal samples.}
    \label{fig:conf-matrix-signal-single}
\end{figure}

To further assess class-level behavior, the signal-level confusion matrix is shown in Fig.~\ref{fig:conf-matrix-signal-single}. The classifier maintained good separation of the normal class, with zero false positives. However, a notable fraction of RBD instances (25\%) were misclassified as ITSC, indicating confusion primarily between fault types. This mirrors findings in other settings and reinforces that, even under limited training exposure, SGDA allows the model to reliably distinguish faults from normality—though refinement is needed to resolve inter-fault ambiguities.

These findings highlight the utility of SGDA for training with limited labeled data but also emphasize the importance of load diversity for multiclass generalization.

\subsection{Evaluation on a Different Motor Configuration}
\label{subsec:other-motor-experiment}

To assess the generalization of the SGDA framework across motor types, we conducted an experiment on a second industrial-grade induction motor (Motor B, RA180L4/2Y3). This motor differs significantly in power, design, and spectral profile compared to the training motor. The goal was to determine whether SGDA can be effectively applied to a new motor setup when only normal-condition data is available from that motor.

In line with the SGDA methodology, the model was retrained using synthetic faults injected into normal current signals from Motor B. No real faulty signals from Motor B were used for training. The evaluation was performed on actual faulty current signals recorded under real ITSC and RBD conditions, as described in Section~\ref{sec:dataset-description}. 

Given the limited dataset,only one signal per class (Normal, ITSC, RBD),evaluation was conducted at the \textit{segment level} using 1-second windows from each test signal. 

\begin{figure}[h!]
    \centering
    \includegraphics[width=0.49\linewidth]{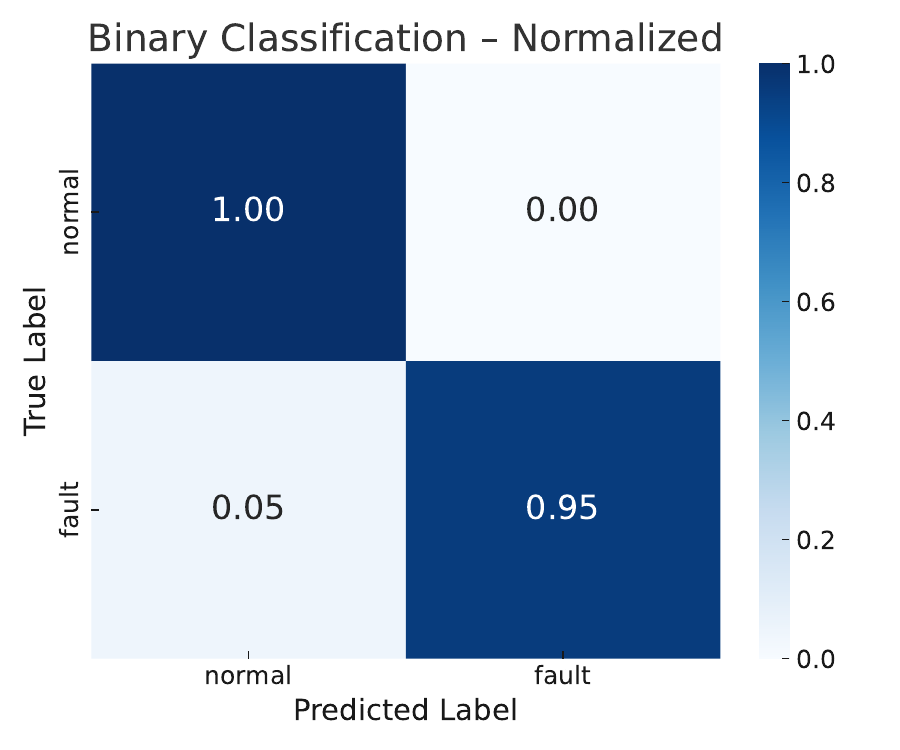}
    \includegraphics[width=0.49\linewidth]{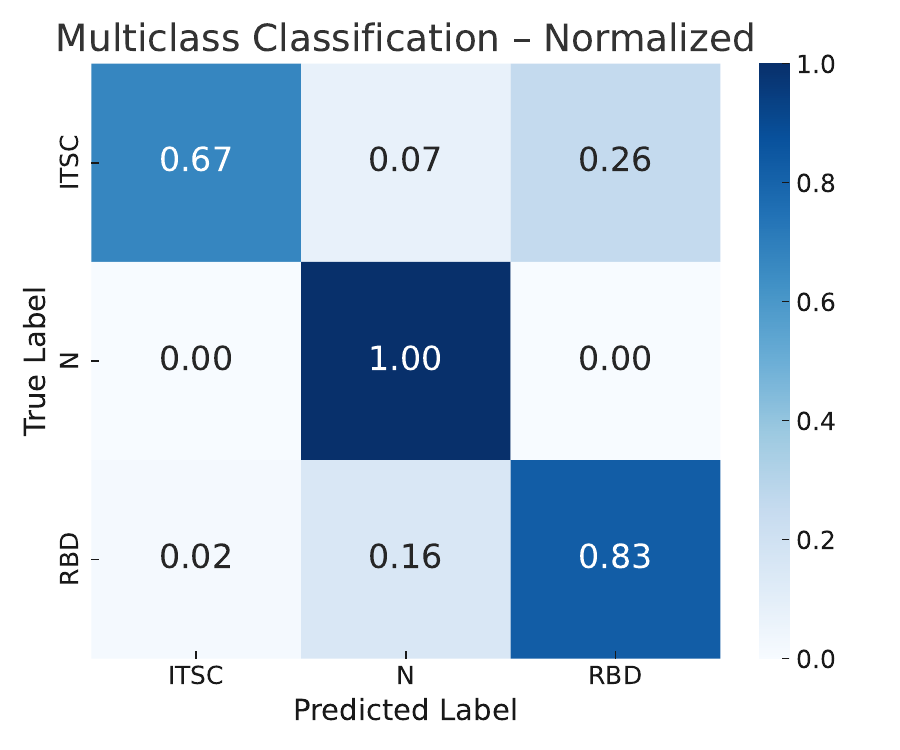}
    \caption{Normalized confusion matrices for segment-level classification on the second motor. Left: Binary classification. Right: Multiclass classification.}
    \label{fig:othermotor_cm}
\end{figure}

Figure~\ref{fig:othermotor_cm} shows the normalized confusion matrices for both binary and multiclass classification.

In the \textbf{binary} case (normal vs. fault), the model achieved a near-perfect separation with 100\% of normal segments correctly classified and 95\% of faulty segments correctly detected. This highlights SGDA’s ability to inject generalizable fault patterns that transfer effectively across motor architectures.

For the \textbf{multiclass} task, the model again performed well, correctly identifying all normal segments and most of the RBD ones (83\%). Some confusion occurred between ITSC and RBD (26\%), which is expected due to their overlapping spectral characteristics. Importantly, no faulty segments were misclassified as normal, reinforcing SGDA's reliability in preserving class separability.

These results demonstrate that SGDA enables effective cross-motor generalization with minimal data—requiring only healthy current signals for training, making it highly suitable for real-world deployment in industrial environments for scalable fault diagnosis.

\section{Impact of Labeled Anomalies on Classification Performance}
\subsection{Experiment: Varying the Number of Real Fault Observations}
To evaluate the necessity of real fault observations for accurate fault classification, we designed a controlled experiment in which we varied the number of real anomalous samples available during training. Using subsets of 2, 4, 8, 10, 20, 50, 100, 200 and 300 labeled anomalies, we trained three models CNN, MLP, and ResNet on both binary and multiclass classification tasks. Performance was assessed on a held-out test set covering the full distribution of motor phases and load conditions.
\subsection{Results and Implications for SGDA}
Tables~\ref{tab:accuracy_combined_all} and~\ref{tab:f1_combined_all} present the accuracy and macro $F_1$ scores across all observation levels. The results reveal a consistent trend, as expected: performance improves with increased access to labeled anomalies, particularly for simpler models like MLP. However, significant variability remains in low-data regimes, and even strong architectures like ResNet require tens of real fault samples to achieve stable results.

In contrast, our proposed SGDA framework achieved high performance without using any real fault data during training. Specifically, SGDA attained an accuracy of 1.00 and $F_1$ of 1.00 for binary classification, and an accuracy of 0.85 and $F_1$ of 0.85 for multiclass classification matching or exceeding the performance of models trained with 100+ labeled examples.

This finding highlights a critical challenge in real-world motor diagnostics, collecting sufficient labeled fault data is difficult and expensive. By leveraging physics-guided synthetic fault injection, SGDA eliminates the need for costly real-world anomaly collection while maintaining a good generalization and diagnostic capability.

\begin{table}[h!]
\centering
\begin{tabular}{llccc}
\toprule
Task & \# of Observations & CNN & MLP & ResNet \\
\midrule
\multirow{8}{*}{\rotatebox[origin=c]{90}{Binary}} 
& 2   & 0.64 $\pm$ 0.08 & 0.63 $\pm$ 0.01 & 0.55 $\pm$ 0.08 \\
& 4   & 0.72 $\pm$ 0.10 & 0.63 $\pm$ 0.00 & 0.67 $\pm$ 0.08 \\
& 8   & 0.79 $\pm$ 0.07 & 0.67 $\pm$ 0.02 & 0.75 $\pm$ 0.08 \\
& 10  & 0.89 $\pm$ 0.03 & 0.66 $\pm$ 0.05 & 0.83 $\pm$ 0.06 \\
& 20  & 0.93 $\pm$ 0.02 & 0.76 $\pm$ 0.02 & 0.73 $\pm$ 0.18 \\
& 50  & 0.97 $\pm$ 0.02 & 0.86 $\pm$ 0.04 & 0.78 $\pm$ 0.10 \\
& 100 & 0.98 $\pm$ 0.02 & 0.89 $\pm$ 0.02 & 0.93 $\pm$ 0.02 \\
& 200 & 1.00 $\pm$ 0.00 & 0.95 $\pm$ 0.00 & 0.98 $\pm$ 0.01 \\
& 300 & 1.00 $\pm$ 0.00 & 0.97 $\pm$ 0.00 & 0.99 $\pm$ 0.00 \\
\midrule
\multirow{8}{*}{\rotatebox[origin=c]{90}{Multiclass}} 
& 2   & 0.45 $\pm$ 0.03 & 0.40 $\pm$ 0.01 & 0.45 $\pm$ 0.03 \\
& 4   & 0.50 $\pm$ 0.02 & 0.44 $\pm$ 0.01 & 0.52 $\pm$ 0.06 \\
& 8   & 0.59 $\pm$ 0.06 & 0.51 $\pm$ 0.05 & 0.62 $\pm$ 0.05 \\
& 10  & 0.62 $\pm$ 0.07 & 0.50 $\pm$ 0.02 & 0.53 $\pm$ 0.10 \\
& 20  & 0.81 $\pm$ 0.03 & 0.63 $\pm$ 0.03 & 0.66 $\pm$ 0.10 \\
& 50  & 0.90 $\pm$ 0.01 & 0.75 $\pm$ 0.04 & 0.74 $\pm$ 0.05 \\
& 100 & 0.96 $\pm$ 0.02 & 0.84 $\pm$ 0.02 & 0.93 $\pm$ 0.03 \\
& 200 & 0.99 $\pm$ 0.01 & 0.90 $\pm$ 0.01 & 1.00 $\pm$ 0.00 \\
& 300 & 1.00 $\pm$ 0.00 & 0.92 $\pm$ 0.01 & 1.00 $\pm$ 0.00 \\
\bottomrule
\end{tabular}
\caption{Accuracy (mean $\pm$ std) for CNN, MLP, and ResNet on binary and multiclass classification tasks across varying numbers of observations.}
\label{tab:accuracy_combined_all}
\end{table}

\begin{table}[h!]
\centering
\begin{tabular}{llccc}
\toprule
Task & \# of Observations & CNN & MLP & ResNet \\
\midrule
\multirow{8}{*}{\rotatebox[origin=c]{90}{Binary}} 
& 2   & 0.56 $\pm$ 0.13 & 0.45 $\pm$ 0.06 & 0.50 $\pm$ 0.09 \\
& 4   & 0.68 $\pm$ 0.17 & 0.48 $\pm$ 0.04 & 0.62 $\pm$ 0.14 \\
& 8   & 0.77 $\pm$ 0.07 & 0.63 $\pm$ 0.04 & 0.74 $\pm$ 0.09 \\
& 10  & 0.88 $\pm$ 0.04 & 0.59 $\pm$ 0.11 & 0.83 $\pm$ 0.06 \\
& 20  & 0.93 $\pm$ 0.02 & 0.75 $\pm$ 0.02 & 0.72 $\pm$ 0.19 \\
& 50  & 0.97 $\pm$ 0.02 & 0.85 $\pm$ 0.04 & 0.78 $\pm$ 0.10 \\
& 100 & 0.98 $\pm$ 0.02 & 0.88 $\pm$ 0.02 & 0.93 $\pm$ 0.02 \\
& 200 & 0.99 $\pm$ 0.00 & 0.95 $\pm$ 0.00 & 0.98 $\pm$ 0.01 \\
& 300 & 1.00 $\pm$ 0.00 & 0.97 $\pm$ 0.00 & 0.99 $\pm$ 0.00 \\
\midrule
\multirow{8}{*}{\rotatebox[origin=c]{90}{Multiclass}} 
& 2   & 0.34 $\pm$ 0.05 & 0.24 $\pm$ 0.03 & 0.34 $\pm$ 0.06 \\
& 4   & 0.44 $\pm$ 0.05 & 0.33 $\pm$ 0.03 & 0.44 $\pm$ 0.09 \\
& 8   & 0.54 $\pm$ 0.08 & 0.41 $\pm$ 0.04 & 0.56 $\pm$ 0.05 \\
& 10  & 0.57 $\pm$ 0.07 & 0.43 $\pm$ 0.04 & 0.45 $\pm$ 0.15 \\
& 20  & 0.80 $\pm$ 0.03 & 0.59 $\pm$ 0.05 & 0.62 $\pm$ 0.11 \\
& 50  & 0.90 $\pm$ 0.01 & 0.72 $\pm$ 0.04 & 0.75 $\pm$ 0.05 \\
& 100 & 0.96 $\pm$ 0.02 & 0.82 $\pm$ 0.02 & 0.93 $\pm$ 0.03 \\
& 200 & 0.99 $\pm$ 0.01 & 0.89 $\pm$ 0.01 & 1.00 $\pm$ 0.00 \\
& 300 & 1.00 $\pm$ 0.00 & 0.92 $\pm$ 0.01 & 1.00 $\pm$ 0.00 \\
\bottomrule
\end{tabular}
\caption{Macro $F_1$-score (mean $\pm$ std) for CNN, MLP, and ResNet on binary and multiclass classification tasks across varying numbers of observations.}
\label{tab:f1_combined_all}
\end{table}

\section{Discussion}

The proposed SGDA framework offers a practical and scalable pathway for implementing fault detection in three-phase induction motors without requiring real fault data during training. For motor owners or operators, the deployment of SGDA involves only minimal setup and domain-specific knowledge. The essential parameters can be retrieved from the motor's nameplate or datasheet. These parameters are used to compute the fault-related frequency bands that guide the synthetic augmentation process.

For data collection, only healthy current signals are required. These signals could be recorded from a single motor phase; this simplifies the acquisition process while still enabling accurate fault detection. A key consideration is the mechanical load level during data collection, which should ideally match the load conditions expected during model deployment. For motors operating continuously in real-time monitoring scenarios, it is recommended to collect training data at the motor’s typical working load. This ensures that the spectral characteristics observed at inference time align with those learned during training.

Alternatively, for motors that are evaluated periodically (e.g., during scheduled maintenance), it is advisable to acquire training signals at full load (100\%). This load level tends to exhibit the most distinct fault-related spectral features, resulting in more robust classifiers. Importantly, the SGDA framework supports training multiple models with minimal overhead. Given that the training is fast and computationally lightweight, users can build separate models tailored for different load levels or use a combined dataset covering all conditions to build a single general-purpose model.

SGDA enables the user to train one or more classifiers—binary (normal vs. fault) and/or multiclass (specific fault types)—based on their operational requirements. Once the models are trained, they can be used in real-time or periodic diagnostics to assess incoming current signals. The model processes the signal in segments and outputs a classification result, indicating whether the motor is operating normally or exhibiting signs of fault.

In summary, SGDA allows engine owners to proactively monitor motor health using only healthy baseline signals and minimal configuration. With the ability to simulate realistic fault conditions and train robust classifiers offline, the approach supports real-time fault diagnosis without the overhead of manual labeling or waiting for rare fault events. This positions SGDA as a practical and cost-effective solution for predictive maintenance and industrial motor monitoring.

\paragraph{Comparison with Existing Augmentation Strategies}

As summarized in Table~\ref{tab:augmentation-comparison}, SGDA uniquely combines the benefits of physics-informed realism, novel fault synthesis, and dynamic augmentation, all without requiring real fault data or computationally expensive simulations. This positions SGDA as a scalable and practical solution for industrial diagnostics, offering adaptability and efficiency compared to prior augmentation strategies.

\begin{table}[h!]
\centering
\renewcommand{\arraystretch}{2}  
{\Huge
\begin{adjustbox}{max width=\textwidth}
\begin{tabular}{lcccccc}
\toprule
 \textbf{Method} &  \textbf{Unsupervised Approach} &  \textbf{No Simulation Needed} &  \textbf{Dynamic Aug.} &  \textbf{Novel Fault Synthesis} &  \textbf{Computational Cost} \\
\midrule
 \textbf{Frequency-Domain Feature Injection~\cite{s22239494,10003718,en17163956}} & \xmark & \cmark & \xmark & \xmark & Low \\
 \textbf{Simulation-Based Augmentation~\cite{e25030414,MA2021115234,s24082575,9547305}} & \cmark & \xmark & \xmark & \cmark & High \\
 \textbf{Domain Transfer/Hybrid Models~\cite{MA2021115234}} & \xmark & \cmark & \cmark & \xmark & High \\
 \textbf{Measured Data Augmentation~\cite{en17163956,jsan13050060}} & \xmark & \cmark & \xmark & \xmark & Low \\
 \textbf{SGDA (Ours)} & \cmark & \cmark & \cmark & \cmark & Low \\
\bottomrule
\end{tabular}
\end{adjustbox}
}

\caption{Comparison of physics-informed data augmentation methods with SGDA across key criteria. The evaluation is based on six criteria: 
(1) \textbf{Unsupervised Approach}—whether real faulty signals are needed for training; 
(2) \textbf{No Simulation Needed}—if fault behaviors are generated via digital twins, FEM, or multiphysics modeling; 
(3) \textbf{Dynamic Augmentation}—whether the method injects variation per training epoch; 
(4) \textbf{Novel Fault Synthesis}—the ability to simulate unseen or hybrid fault behaviors; 
(5) \textbf{Computational Cost}—training or simulation overhead;}
\label{tab:augmentation-comparison}
\end{table}

\subsection{Handling Partial or Unknown Motor Parameters}

Although SGDA leverages known motor parameters to calculate defect-specific frequencies, many practical scenarios involve incomplete or uncertain specifications. In such cases, one can estimate slip or other key parameters by combining nameplate data (e.g., rated speed, voltage, power) with short test runs under nominal load conditions. By measuring the main operational frequency and approximate rotor speed, the slip can be inferred, and typical sideband offsets for rotor and stator defects can be calculated within a plausible range (e.g., 1–5\% slip).  

Crucially, the SGDA framework accommodates parameter uncertainty \textit{natively} through its synthetic augmentation mechanism. The key is the effective frequency window $\epsilon_f$, which defines a small band around each expected defect frequency where Gaussian-shaped peaks are injected. Rather than requiring precise values of slip or other parameters, SGDA injects spectral anomalies within a controlled neighborhood around the expected frequency. For instance, rotor bar faults typically induce components near $f \pm 2s f_0$, but in the absence of exact slip values, $\epsilon_f$ allows SGDA to simulate a distribution of possible fault manifestations.

This localized injection ensures that the model is exposed to a realistic spread of spectral behavior, capturing the imprecision inherent in real-world diagnostics. The value of $\epsilon_f$ thus acts as a tunable proxy for parameter uncertainty: smaller values are suitable when motor parameters are well known, while larger windows improve robustness when only partial or approximate information is available.

These strategies ensure SGDA is not strictly limited to motors with fully documented parameters, extending its applicability to a wide range of industrial setup, where full specifications may be unknown or unreliable.

\subsection{Limitations}

\begin{itemize}
    \item \textbf{Overlapping Fault Features:} 
    \textit{Mitigation Strategies:} Our current approach might be extended by injecting a broader range of fault types to better distinguish overlapping features. However, this possibility has not yet been experimentally verified.
    \item \textbf{Scalability to Different Motor Types:} 
    Our approach has been tested on two real engines configurations under different faults and operational conditions. The performance and generalization capability to other motor types remain to be evaluated, which is a key area for future work.
    \item \textbf{Predefined Engine Configurations:} 
    The method currently relies on predefined engine configurations. This limitation can be addressed by incorporating a larger variety of engine types. A promising direction is the integration of digital twins that simulate faults and different operating conditions. This approach could generate extensive training pairs (engine configuration, observed signal such as current, vibration, etc.), facilitating the development of more robust algorithms that integrate multiple data sources in the decision-making process.
    \item \textbf{Parameter Availability:} 
    It is important to note that most engine parameters required for MCSA, and hence for our method, are typically provided on the motor’s documentation. If not available, these parameters can usually be determined experimentally.
    \item \textbf{Fault Severity Estimation and Prognostics}: Detecting a fault is the first step; assessing its severity and predicting remaining useful life is the next. Operators need to know how bad a fault is – is it a minor imbalance or a severe misalignment that will cause failure in days? Physics-informed modeling combined with AI could enable more nuanced diagnostics. For example, one could use the motor’s physical model to simulate progression of a crack in the rotor, generating data for different severity levels, and train a model to classify the degree of fault. Future research should integrate fault detection with fault diagnosis (identifying the fault type and location) and prognostics (predicting fault evolution).
\end{itemize}

\section{Conclusion}

In this work, we introduced SGDA—an unsupervised, physics-informed framework for fault detection in three-phase induction motors. SGDA requires only healthy motor current signals and basic motor parameters to generate realistic, fault-annotated training data in the frequency domain. This eliminates the need for collecting costly and rare real-world fault samples while maintaining high classification performance.

We demonstrated the flexibility and robustness of SGDA through extensive experiments. The system was tested across different operational settings, including full and partial load conditions, and validated on unseen motor phases. We also showed that SGDA generalizes to an entirely different motor configuration, confirming its transferability.

For classifier integration, we evaluated several models and adopted a ResNet-based architecture for its strong performance in both binary and multiclass scenarios. The end-to-end SGDA pipeline—from healthy signal collection to fault classification—is lightweight, easy to deploy, and adaptable to various use cases including real-time monitoring and periodic diagnostics.

SGDA presents a practical and scalable solution for deploying intelligent motor diagnostics in industrial settings—without the burden of fault data collection or manual labeling.


\section*{Acknowledgments}
This work was supported by the Ministry of Economic Development of the Russian Federation (code 25-139-66879-1-0003). The computation for this research was performed using the computational resources of HPC facilities at HSE University. The authors are grateful to LIMAN LLC for the data provided.

\clearpage 
\bibliographystyle{elsarticle-num}
\bibliography{references}

\appendix

\section{MCSA-based Fault Frequency Extraction Formulas}
\label{appendix:mcsa}

\begin{itemize}
    \item \textbf{Rotor Bar Defect Frequencies:}
    \[
    f = f_1 \left(1 \pm 2n s\right),
    \]
    where $f_1$ is a supply frequency and $s$ is a motor slip.
    \item \textbf{Inter-Turn Short Circuit (ITSC) Frequencies:}
    \[
    f_{\text{ITSC}} = k f_s \mp m f_r
    \]
    where $f_s$ is the supply frequency, $f_r$ is the rotor (rotation) frequency, \(k\) is an odd harmonic multiplier (e.g., \(1, 3, 5, \dots\)) and \(m\) is an integer representing the sideband order.

    \item \textbf{Bearing Defect Frequencies:} \\
    For a rolling element bearing:
    \begin{itemize}
        \item \textbf{Outer Race (BPFO):}
        \[
        \text{BPFO} = \frac{n}{2} f_r \left(1 - \frac{D_{ball}}{D_{pit}} \cos\beta\right)
        \]
        \item \textbf{Inner Race (BPFI):}
        \[
        \text{BPFI} = \frac{n}{2} f_r \left(1 + \frac{D_{ball}}{D_{pit}} \cos\beta\right)
        \]
        \item \textbf{Rolling Element (BSF):}
        \[
        \text{BSF} = \frac{D_{pit}}{2D_{ball}} f_r \left(1 - \left(\frac{D_{ball}}{D_{pit}} \cos\beta\right)^2\right)
        \]
    \end{itemize}
    where $n$ is a number of rolling elements, $D_{pit}$ is a pitch (mean) diameter of the bearing (m), $D_{ball}$ is a diameter of the rolling element (m), $f_r$ is a shaft (rotational) frequency (Hz), $\beta$ is a contact angle of the rolling elements (radians).
\end{itemize}

\newpage
\section{Experiments with Various Hyperparameters}
\label{appendix:experiments}

\subsection{Batch Size}

\begin{figure}[H]
    \centering
    \includegraphics[width=0.85\linewidth]{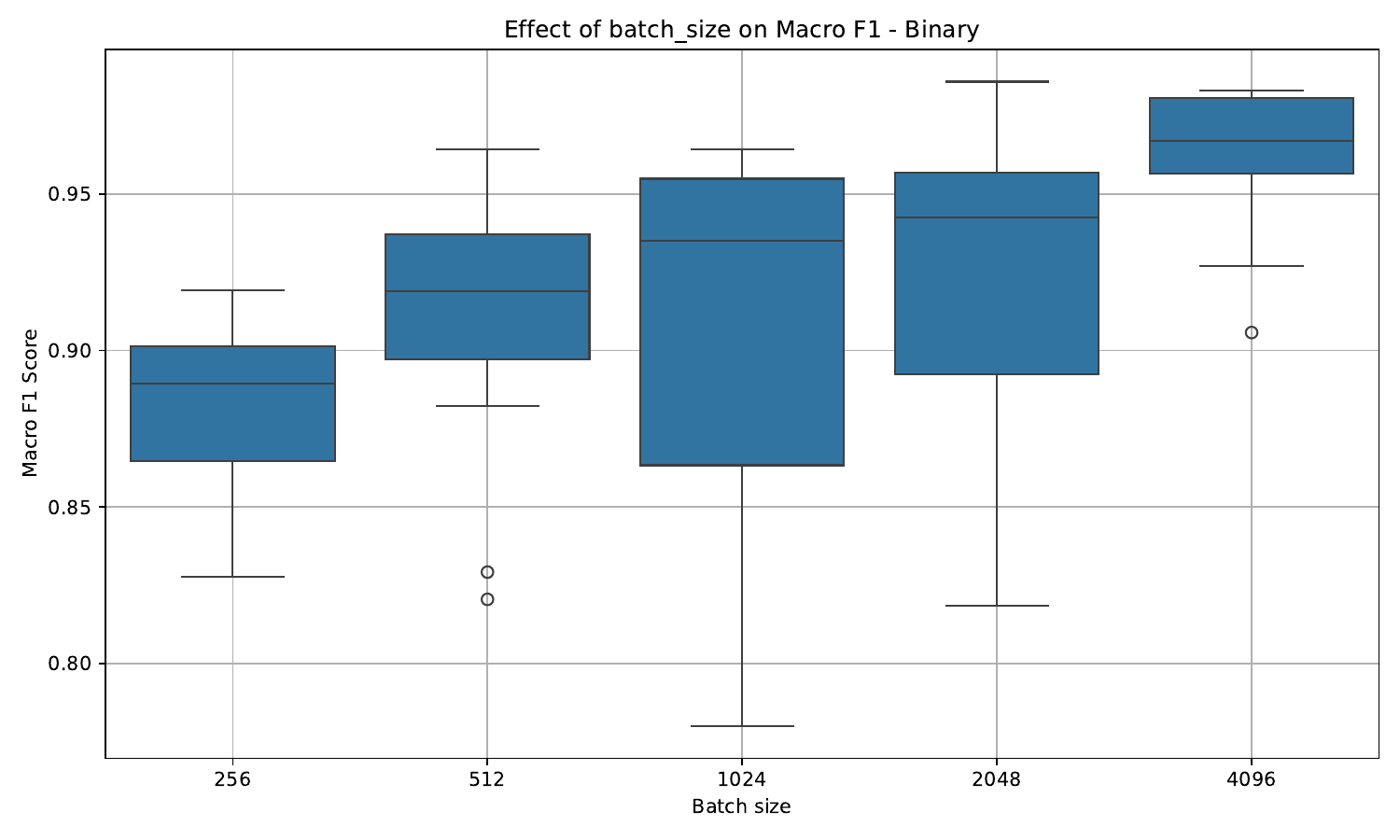}
    \caption{Effects of batch size on Macro $F_1$ in the binary classification.}
    \label{fig:analysis:batch_binary}
\end{figure}

\begin{figure}[H]
    \centering
    \includegraphics[width=0.85\linewidth]{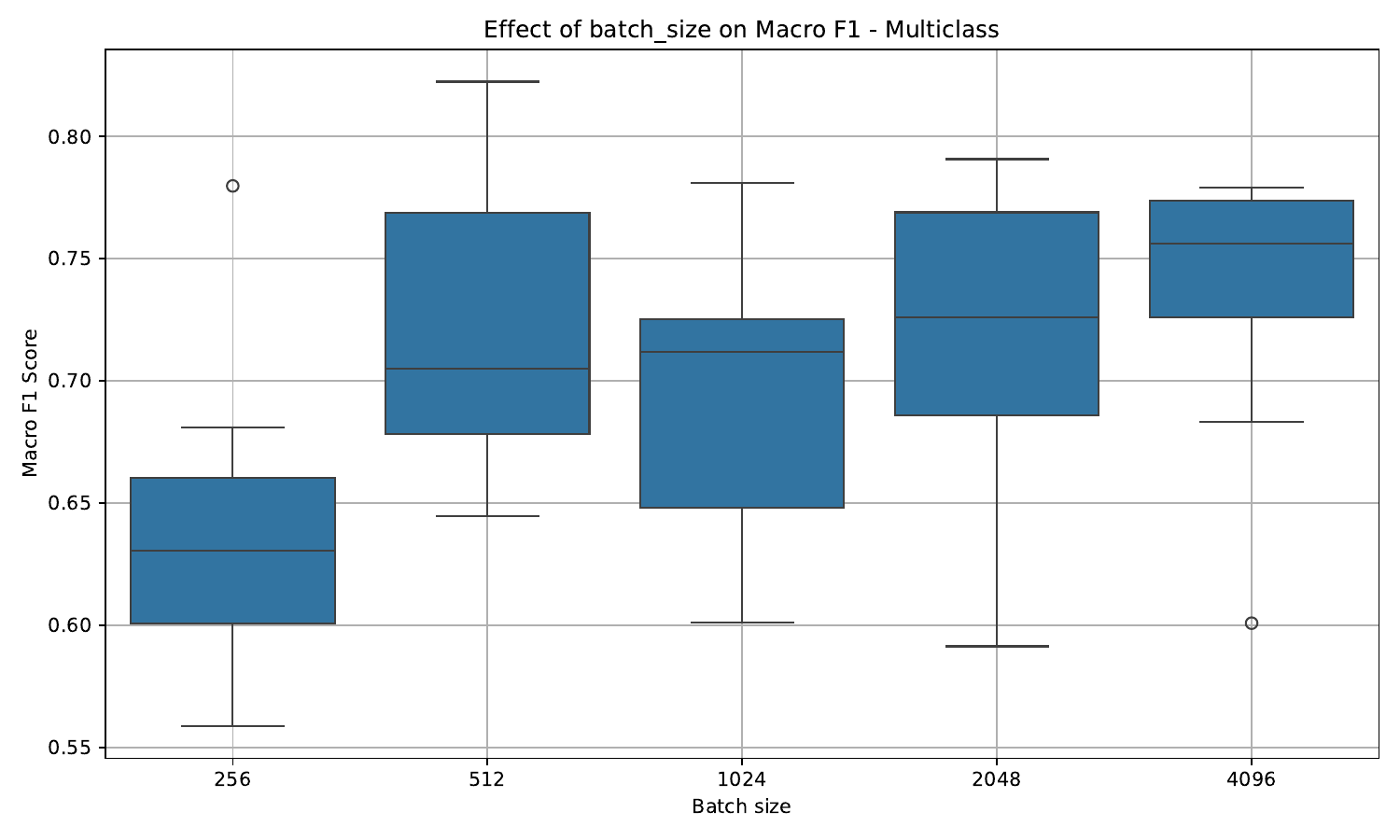}
    \caption{Effects of batch size on Macro $F_1$ in the multiclass classification.}
    \label{fig:analysis:batch_multiclass}
\end{figure}

\subsection{Normalization Mode}

Normalization modes described in \autoref{subsection:rsa}:
\begin{itemize}
    \item Individual segment–channel normalization (\texttt{per}),
    \item Global per-channel normalization (\texttt{global}).
\end{itemize}

\begin{figure}[H]
    \centering
    \includegraphics[width=0.8\linewidth]{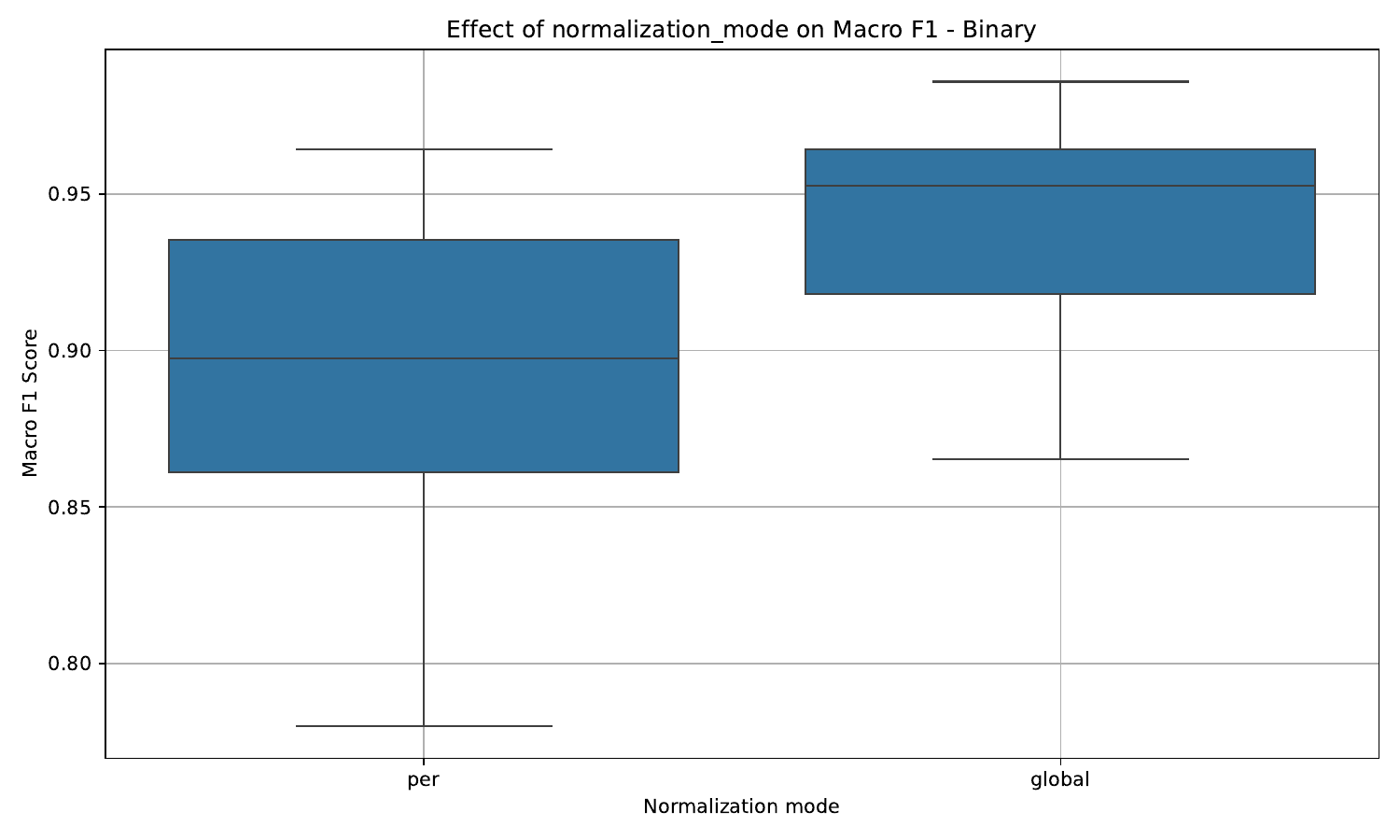}
    \caption{Effects of normalization mode on Macro $F_1$ in the binary classification.}
    \label{fig:analysis:normalization_binary}
\end{figure}

\begin{figure}[H]
    \centering
    \includegraphics[width=0.8\linewidth]{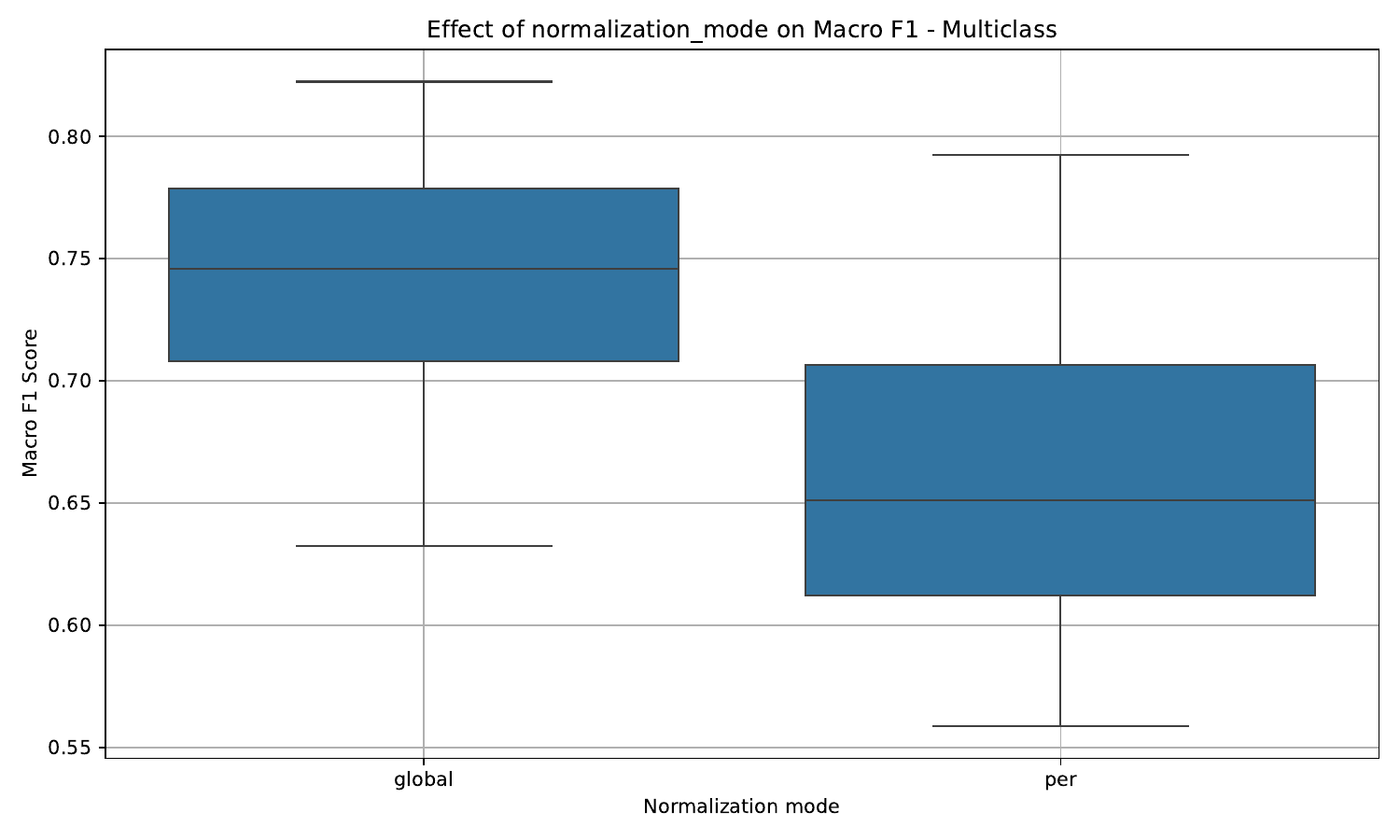}
    \caption{Effects of normalization mode on Macro $F_1$ in the multiclass classification.}
    \label{fig:analysis:normalization_multiclass}
\end{figure}

\subsection{Dropout}

\begin{figure}[H]
    \centering
    \includegraphics[width=0.85\linewidth]{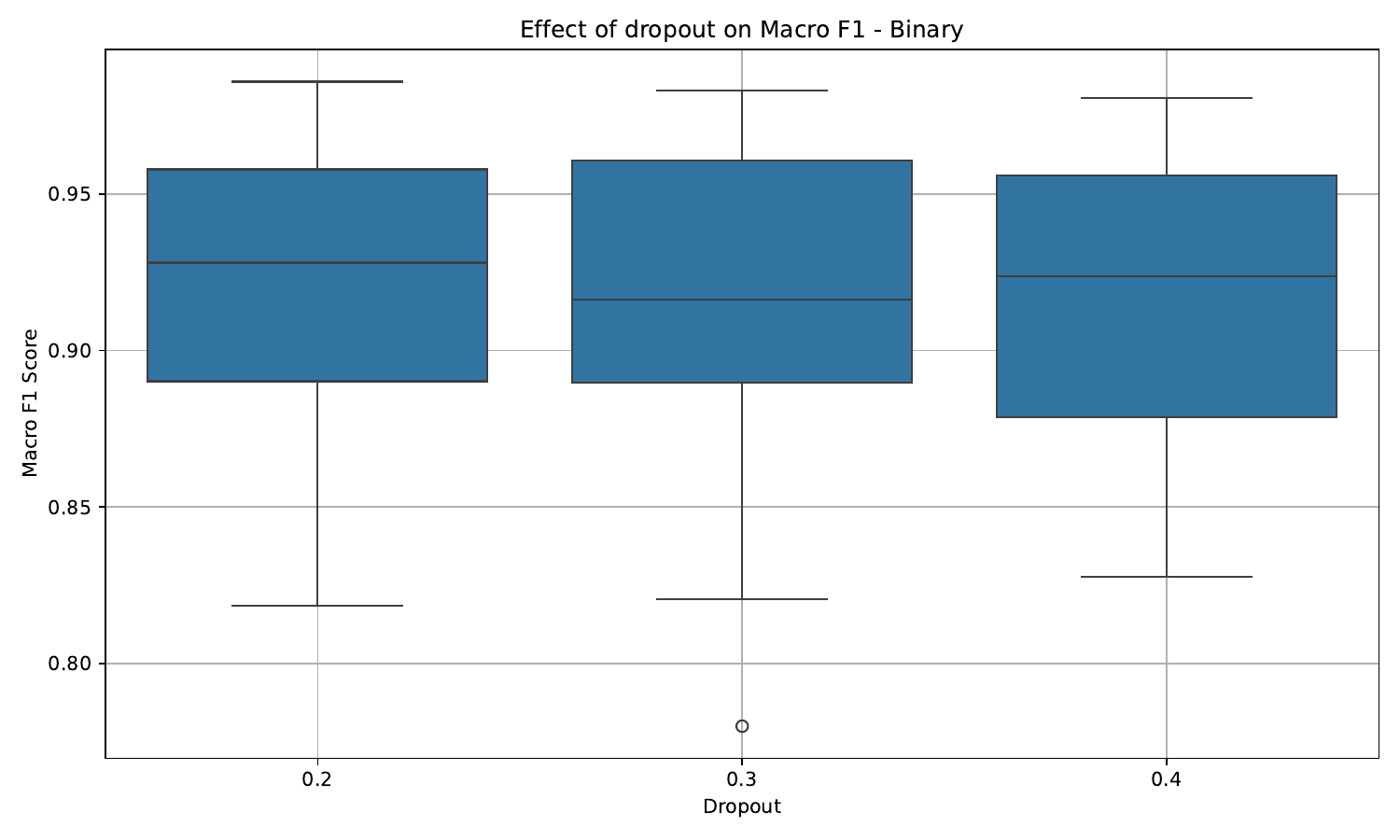}
    \caption{Effects of dropout value on Macro $F_1$ in the binary classification.}
    \label{fig:analysis:dropout_binary}
\end{figure}

\begin{figure}[H]
    \centering
    \includegraphics[width=0.85\linewidth]{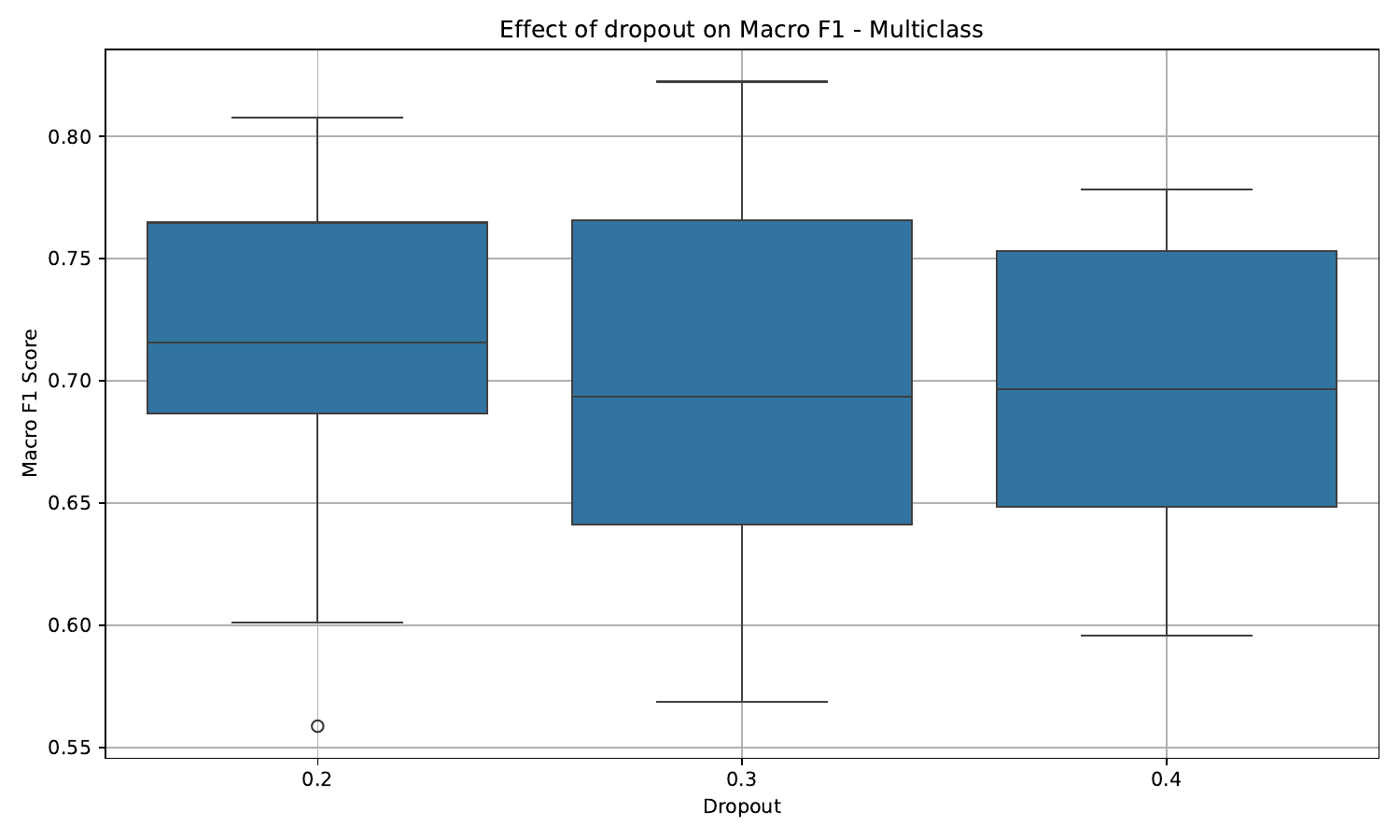}
    \caption{Effects of dropout value on Macro $F_1$ in the multiclass classification.}
    \label{fig:analysis:dropout_multiclass}
\end{figure}

\subsection{Attention Mask}

\textbf{Attention Mask} is a fixed, domain-informed binary tensor that identifies \textit{diagnostic frequency bands} corresponding to fault characteristic frequencies $\pm$ an effective window $\epsilon_f$. The mask assigns a value of 1 to frequency bins within these fault-centered bands and 0 elsewhere. It enables selective modulation of input features by allowing the model to apply distinct learnable weights to fault-relevant and non-relevant spectral regions, thereby emphasizing fault signatures while preserving end-to-end differentiability.

\begin{figure}[H]
    \centering
    \includegraphics[width=0.75\linewidth]{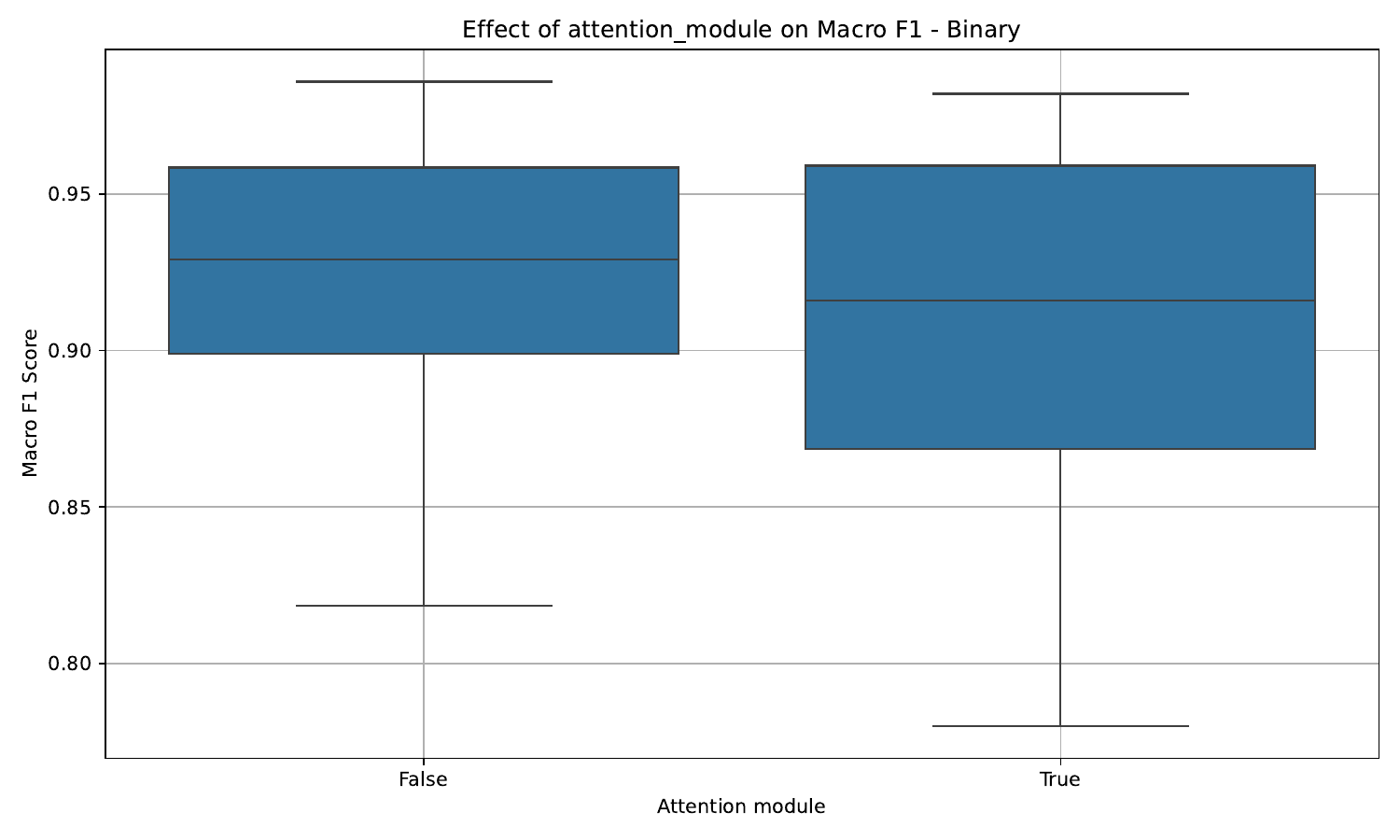}
    \caption{Effects of attention module on Macro $F_1$ in the binary classification.}
    \label{fig:analysis:attention_binary}
\end{figure}

\begin{figure}[H]
    \centering
    \includegraphics[width=0.75\linewidth]{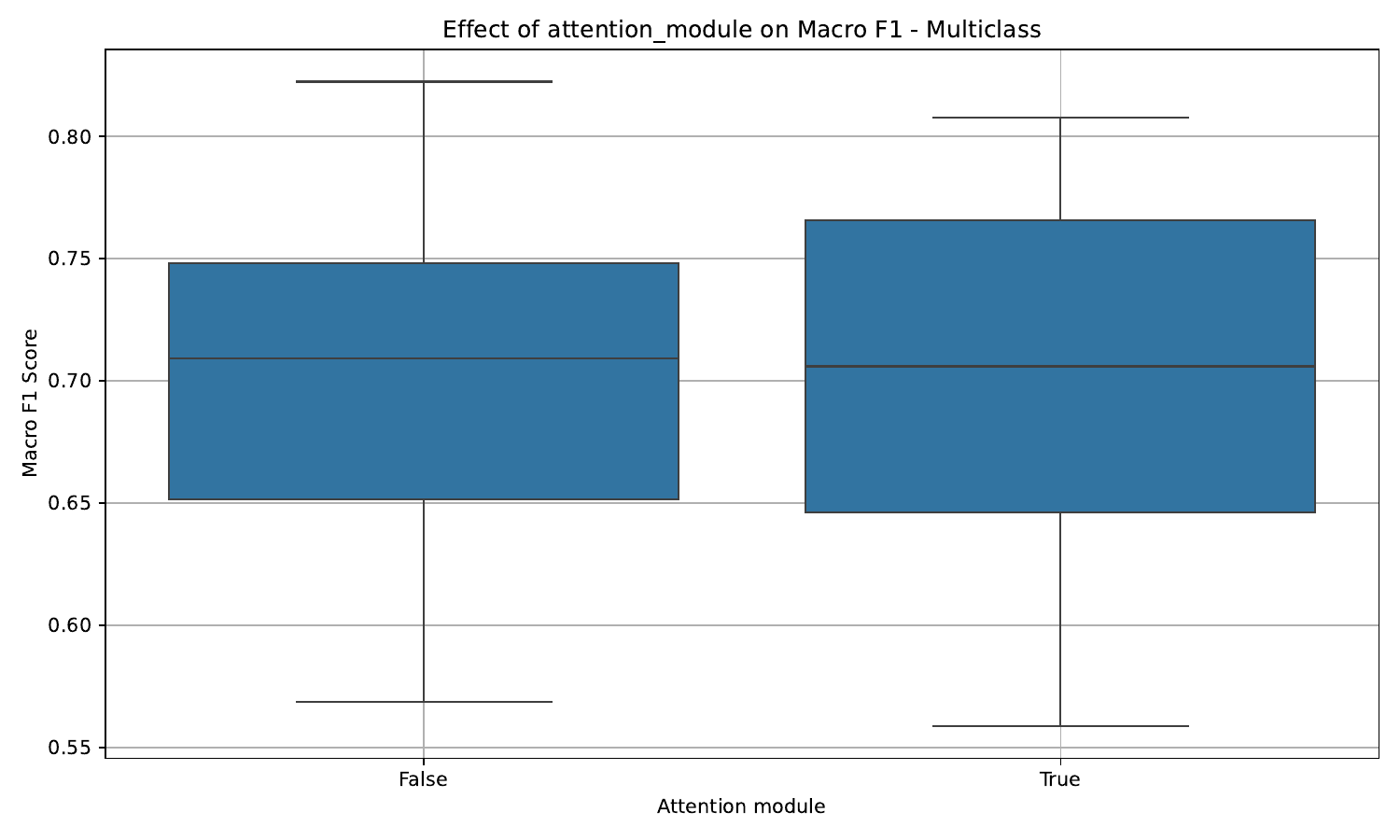}
    \caption{Effects of attention module on Macro $F_1$ in the multiclass classification.}
    \label{fig:analysis:attention_multiclass}
\end{figure}

\end{document}